\documentclass[10pt,journal,compsoc]{IEEEtran}
\usepackage{cite}
\usepackage{framed,multirow}
\usepackage{amssymb}
\usepackage{latexsym}
\usepackage{url}
\usepackage{xcolor}
\usepackage{hyperref}
\definecolor{newcolor}{rgb}{.8,.349,.1}
\usepackage{tabularx}
\usepackage{threeparttable}
\usepackage{booktabs}
\usepackage{graphicx}
\usepackage[switch,columnwise]{lineno}
\usepackage{amsmath}
\usepackage{array}
\usepackage{verbatim}
\usepackage{soul}
\usepackage{colortbl}
\usepackage{enumitem}

\def\etal{{\em et al.}}
\def\eg{{\em e.g.}}
\def\ie{{\em i.e.}}

\begin{document}

\title{Source-Free Unsupervised Domain Adaptation: A Survey}

\author{Yuqi Fang, 
Pew-Thian Yap,
Weili Lin,  
Hongtu Zhu, and 
Mingxia Liu
\IEEEcompsocitemizethanks{\IEEEcompsocthanksitem Y. Fang, P.-T. Yap, W. Lin and M. Liu are with the Department of Radiology and Biomedical Research Imaging Center, University of North Carolina at Chapel Hill, Chapel Hill, NC 27599, United States. H.~Zhu is with the Department of Biostatistics and Biomedical Research Imaging Center, University of North Carolina at Chapel Hill, NC 27599, USA. 
Corresponding author: M.~Liu  (mxliu@med.unc.edu).
}
}
 
\markboth{Source-Free Unsupervised Domain Adaptation: A Survey}
{Fang \MakeLowercase{\textit{et al.}}: Source-Free Unsupervised Domain Adaptation: A Survey}

\IEEEtitleabstractindextext{
\begin{abstract}
Unsupervised domain adaptation (UDA) via deep learning has attracted appealing attention for tackling domain-shift problems caused by distribution discrepancy across different domains.
Existing UDA approaches highly depend on the accessibility of source domain data, which is usually limited in practical scenarios due to privacy protection, data storage and transmission cost, and computation burden.
To tackle this issue, many source-free unsupervised domain adaptation (SFUDA) methods have been proposed recently, which perform knowledge transfer from a pre-trained source model to unlabeled target domain with source data inaccessible. 
A comprehensive review of these works on SFUDA is of great significance. 
In this paper, we provide a timely and systematic literature review of existing SFUDA approaches from a technical perspective. 
Specifically, we categorize current SFUDA studies into two groups, \ie, white-box SFUDA and black-box SFUDA, and further divide them into finer subcategories based on different learning strategies they use.
We also investigate the challenges of methods in each subcategory, discuss the advantages/disadvantages of white-box and black-box SFUDA methods, conclude the commonly used benchmark datasets, and summarize the popular techniques for improved generalizability of models learned without using source data. 
We finally discuss several promising future directions in this field. 
\end{abstract}

}

\maketitle

\begin{figure*}[!t]
    \setlength{\abovecaptionskip}{0pt}
    \setlength{\belowcaptionskip}{-2pt}
    \setlength{\abovedisplayskip}{-2pt}
    \setlength{\belowdisplayskip}{-2pt}
	\centering
	\includegraphics[width=\textwidth]{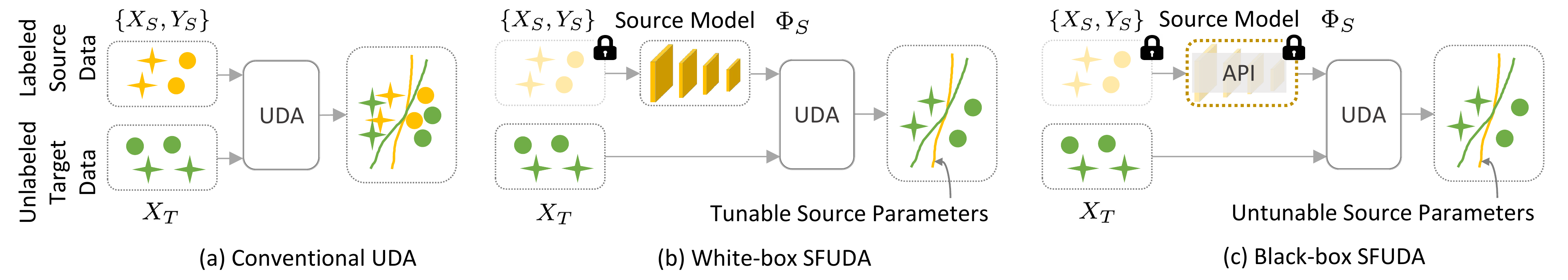}
	\caption{Illustration of (a) conventional unsupervised domain adaptation (UDA), (b) white-box source-free UDA (SFUDA), and (c) black-box SFUDA. 
	Compared with (a) conventional UDA that relies on labeled source data $\{X_S, Y_S\}$ and unlabeled target data $X_T$, (b, c) SFUDA performs knowledge transfer by directly leveraging a pre-trained source model $\Phi_S$ and unlabeled target data $X_T$. The difference between (b) white-box SFUDA and (c) black-box SFUDA lies in whether the learnable parameters of the source model $\Phi_S$ are accessible or not.
	API: application programming interface.
 }
	\label{fig_compare_UDA_SFUDA}
\end{figure*}

\IEEEraisesectionheading{\section{Introduction}\label{sec_introduction}}
\IEEEPARstart{D}{eep} learning, based on deep neural networks with representation learning, has emerged as a promising technique and made remarkable progress over the past decade, covering the field of computer vision~\cite{voulodimos2018deep, hassaballah2020deep}, medical data analysis~\cite{shendg2017deep, litjens2017survey}, natural language processing~\cite{otter2020survey, young2018recent}, etc.
For problems with multiple domains (\eg, different datasets or imaging sites), the typical learning process of a deep neural network is to transfer the model learned on a source domain to a target domain. 
However, performance degradation is often observed when there exists a distribution gap between the source and target domains, which is termed ``domain shift'' problem~\cite{li2017deeperDG,sankaranarayanan2018learning,zhou2022domain}.
To tackle this problem, various domain adaptation algorithms~\cite{wang2018deepsurvey, guan2021domain} have been proposed to perform knowledge transfer by reducing inter-domain distribution discrepancy.
To avoid intensive burden of data annotation, unsupervised domain adaptation has achieved much progress~\cite{dong2021and, ganin2015unsupervised, saito2018maximum, fang2022unsupervised}. 
As illustrated in Fig.~\ref{fig_compare_UDA_SFUDA}~(a), unsupervised domain adaptation aims to transfer knowledge from a labeled source domain to a target domain without accessing any target label information. 

Existing deep learning studies on unsupervised domain adaptation highly depend on the accessibility of source data, which is usually limited in practical scenarios due to the following possible reasons. 
(1) \emph{Data privacy protection}. Many source datasets containing confidential information, such as medical and facial data, are not available to third parties due to 
privacy and security protection.
(2) \emph{Data storage and transmission cost}. The storage and transmission of large-scale source datasets, such as ImageNet~\cite{deng2009imagenet}, could bring much economic burden. 
(3) \emph{Computation burden}. Training on extremely large source datasets requires high computational resources, which is not practical, especially in real-time deployment cases. 
Thus, there is a high demand for source-free unsupervised domain adaptation (SFUDA) methods that 
transfer a pre-trained source model to the unlabeled target domain \emph{without accessing any source data}~\cite{nayak2021mining, kundu2020universal, huang2021model, yang2021exploiting}. 

Many promising SFUDA algorithms have been developed recently to address problems in the fields of object recognition~\cite{kundu2022concurrentsub}, semantic segmentation~\cite{liu2021sourceseman}, image classification~\cite{chen2021self}, object detection~\cite{saltori2020sf}, face anti-spoofing~\cite{liu2022sourceCon}, etc.
A comprehensive review of current studies on SFUDA as well as an outlook on future research directions are urgently needed.
Liu~\etal~\cite{liu2021data} present a review on data-free knowledge transfer, where SFUDA only accounts for part of the review and the taxonomy of SFUDA is generally rough. 
And a large number of relevant studies have emerged in the past year, but the related papers are not included in that survey.
In addition, their work does not cover commonly used datasets in this research field. 

\begin{figure*}[!t]
\setlength{\abovecaptionskip}{0pt}
\setlength{\belowcaptionskip}{-2pt}
\setlength{\abovedisplayskip}{-2pt}
\setlength{\belowdisplayskip}{-2pt}
	\centering
	\includegraphics[width=0.98\textwidth]{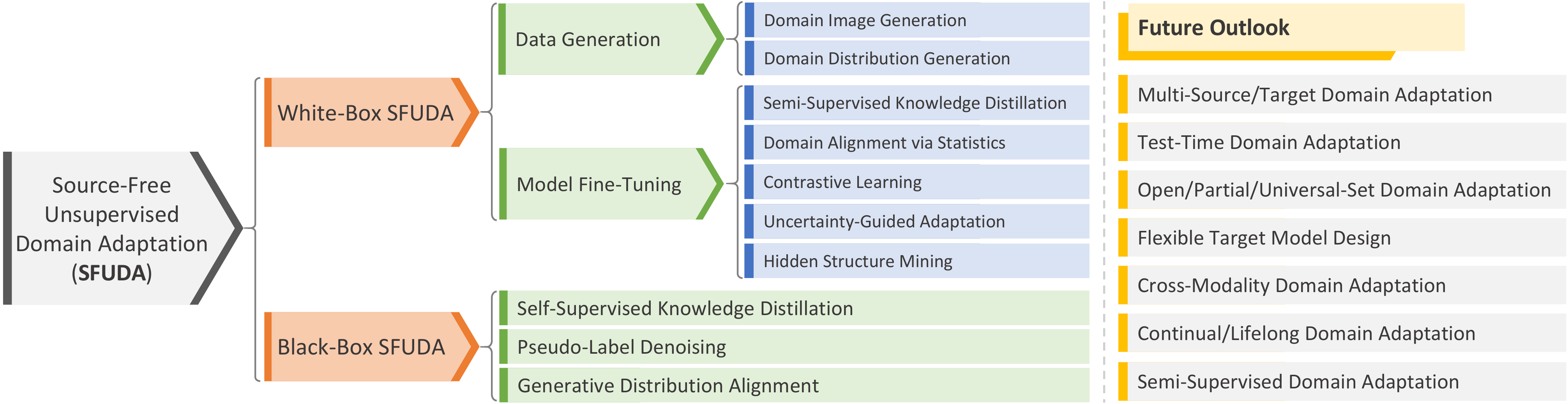}
	\caption{Taxonomy of existing source-free unsupervised domain adaptation (SFUDA) methods, as well as future outlook.}
	\label{fig_category_framework}
\end{figure*}

To fill the gap, in this paper, we provide a timely and thorough literature review of existing deep learning studies on source-free unsupervised domain adaptation.
Our goal is to cover SFUDA studies of the past few years and provide a detailed and systematic SFUDA taxonomy. 
Specifically, we classify existing SFUDA approaches into two broad categories: (1) \emph{white-box SFUDA} as shown in Fig.~\ref{fig_compare_UDA_SFUDA}~(b) and (2) \emph{black-box SFUDA} as illustrated in Fig.~\ref{fig_compare_UDA_SFUDA}~(c). 
The difference between them lies in whether the model parameters of the pre-trained source model are available or not.
Based on different learning strategies they use, we further subdivide white-box and black-box SFUDA methods into finer categories, and the overall taxonomy is shown in Fig.~\ref{fig_category_framework}. 
Moreover, we discuss the challenges and insight for methods in each category, provide a comprehensive comparison between white-box and black-box SFUDA approaches, summarize commonly used datasets in this field as well as popular techniques to improve model generalizability across different domains.
We have to point out that SFUDA is still under vigorous development, so we further discuss the main challenges and provide insights into potential future directions accordingly.

\if false
\begin{figure*}[!t]
\setlength{\abovecaptionskip}{0pt}
\setlength{\belowcaptionskip}{-2pt}
\setlength{\abovedisplayskip}{-2pt}
\setlength{\belowdisplayskip}{-2pt}
	\centering
	\includegraphics[width=0.98\textwidth]{figures/fig_category_framework_v2.pdf}
	\caption{The taxonomy of the existing source-free unsupervised domain adaptation (SFUDA) methods, as well as challenges and future outlooks.}
	\label{fig_category_framework}
\end{figure*}
\fi

The rest of this survey is organized as follows.  
Section~\ref{sec_white_SFUDA} and Section~\ref{sec_black_SFUDA} review existing white-box and black-box SFUDA methods, respectively.
In Section~\ref{sec_discussion}, we compare white-box and black-box SFUDA and present useful strategies to improve model generalization.
Section~\ref{sec_challenge_futurework} discusses challenges of existing studies and future research directions. 
Finally, we conclude this paper in Section~\ref{sec_conclusion}.

\if false
\section{Preliminaries}\label{sec_preliminaries}
In this section, we give the mathematical formulation of source-free unsupervised domain adaptation.
Denote $\Phi_S$ as the source model well-trained based on the labeled source domain $\{X_S, Y_S\}$, and denote $X_T$ as the unlabeled target domain.
The goal of SFUDA is to learn a target model $\Phi_T$ for improved target inference with only the pre-trained source model $\Phi_S$ and unlabeled target domain $X_T$.
\fi

\section{White-Box Source-Free Unsupervised Domain Adaptation}\label{sec_white_SFUDA}
Denote $\Phi_S$ as the source model well-trained based on the labeled source domain $\{X_S, Y_S\}$, where  $X_S$ and $Y_S$ represent source data and the corresponding label information, respectively. 
Denote $\{X_T$\} as the unlabeled target domain with only target samples $X_T$. 
The goal of SFUDA is to learn a target model $\Phi_T$ for improved target inference based on the pre-trained source model $\Phi_S$ and unlabeled target data $X_T$.
\emph{In the setting of white-box source-free domain adaptation, the source data (}\ie, \emph{$X_S$ and $Y_S$) cannot be accessed but the training parameters of the source model $\Phi_S$ are available}.
As shown in the upper middle of Fig.~\ref{fig_category_framework}, existing white-box SFUDA studies can be divided into two categories: \textbf{Data Generation Method} and 
\textbf{Model Fine-Tuning Method}, with details elaborated as follows.

\subsection{Data Generation Method}\label{sec_data_generation}
\subsubsection{Domain Image Generation}\label{sec_domain_image_generation}
\if false
\begin{figure}[!t]
\setlength{\abovecaptionskip}{0pt}
\setlength{\belowcaptionskip}{-2pt}
\setlength{\abovedisplayskip}{-2pt}
\setlength{\belowdisplayskip}{-2pt}
	\centering
	\includegraphics[width=0.49\textwidth]{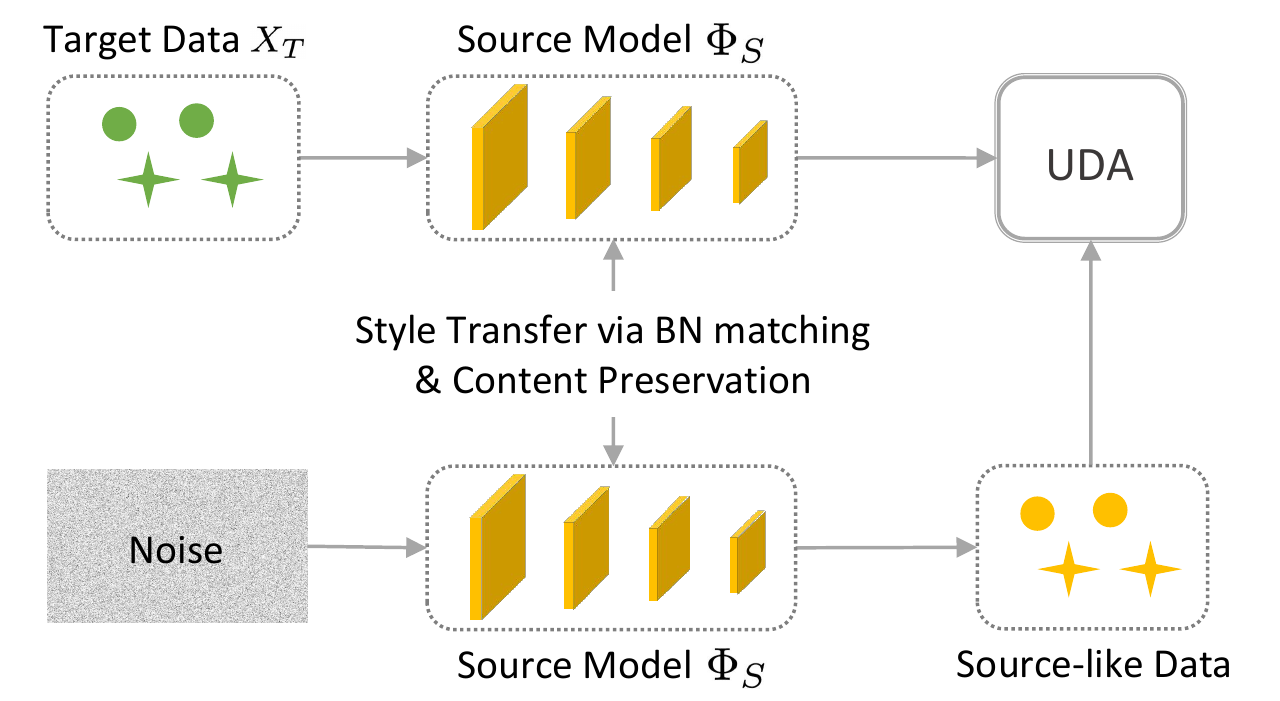}
	\caption{
	The illustration of \emph{Batch Normalization Statistics Transfer} methods for source image generation in source-free unsupervised domain adaptation.
	By matching batch normalization (BN) statistics between the upper and lower branches, the source-like data can be generated, which preserve the target content but with source style.
	Then, the unsupervised domain adaptation can be performed between source-like data and target data.
	}
	\label{fig_ImageGen_1BNtransfer}
\end{figure}
\fi

Many studies aim to generate source-like image data and achieve cross-domain adaptation by readily applying standard unsupervised domain adaptation techniques.
Based on different image generation strategies, these studies can be divided into the following three subcategories: (1) batch normalization statistics transfer, (2) surrogate source data construction, and (3) generative adversarial network (GAN) based image generation.

(\textbf{1}) \textbf{Batch Normalization Statistics Transfer}. 
Considering that batch normalization (BN) stores the running mean and variance for a mini-batch of training data in each layer of a deep learning model, some studies~\cite{yang2022sourcefourier, hou2020source, hong2022source} explicitly leverage such BN statistics for image style transfer, as illustrated in Fig.~\ref{fig_ImageGen_1BNtransfer}.
For instance, Yang~\etal~\cite{yang2022sourcefourier} generate source-like images via a two-stage coarse-to-fine learning strategy. 
In the coarse image generation step, BN statistics stored in the source model are leveraged to preserve the \emph{style characteristics} of source images and also maintain the content information of target data.
In the fine image generation step, an image generator based on Fourier Transform is developed to remove ambiguous textural components of generated images and further improve image quality.
With generated source-like images and given target images, a contrast distillation module and a compact consistency measurement module are designed to perform feature-level and output-level adaptation, respectively.
Similarly, Hou~\etal~\cite{hou2020source} perform \emph{style transfer} by matching BN statistics of generated source-style image features with those saved in the pre-trained source model for image translation.
Hong~\etal~\cite{hong2022source} generate source-like images by designing a \emph{style-compensation transformation} architecture guided by BN statistics stored in the source model and the generated reliable target pseudo-labels.

\begin{figure}[!t]
\setlength{\abovecaptionskip}{0pt}
\setlength{\belowcaptionskip}{-2pt}
\setlength{\abovedisplayskip}{-2pt}
\setlength{\belowdisplayskip}{-2pt}
	\centering
	\includegraphics[width=0.49\textwidth]{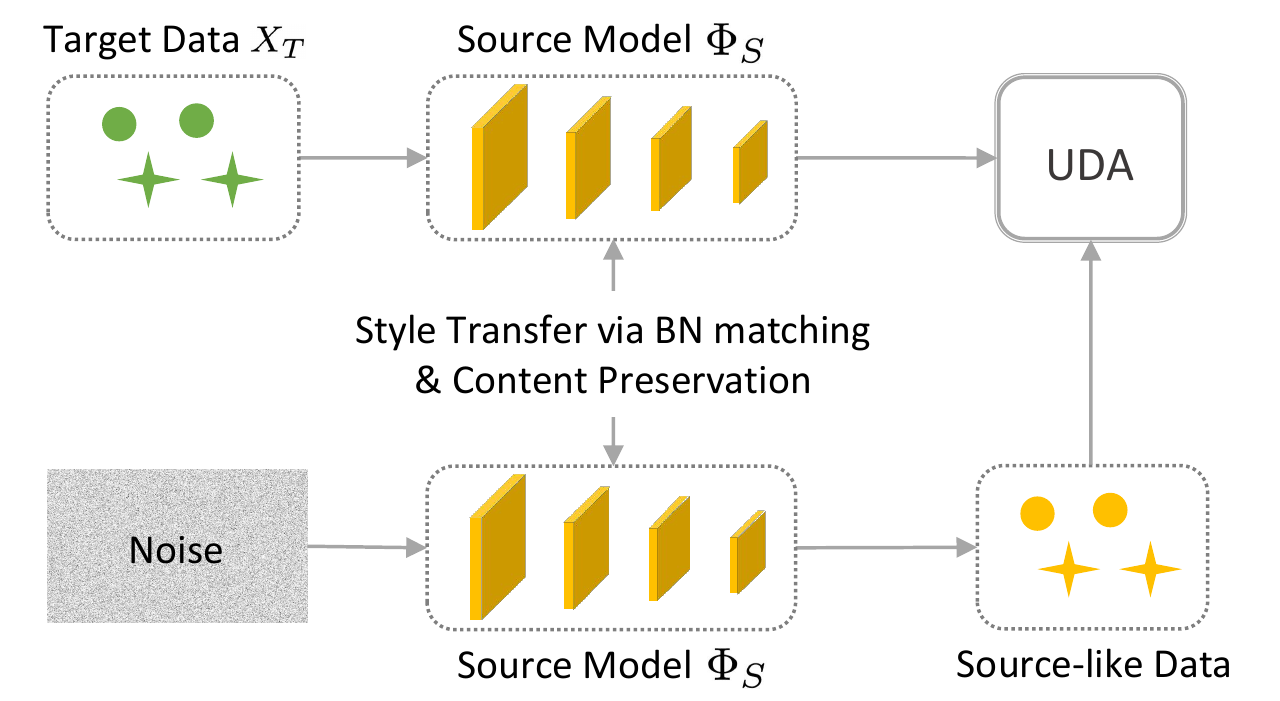}
	\caption{
	Illustration of \emph{Batch Normalization Statistics Transfer} methods for source image generation. 
	By matching batch normalization (BN) statistics between the upper and lower branches, source-like data can be generated by preserving the target content but with source style. 
	Unsupervised domain adaptation (UDA) can then be performed between source-like data and target data.
	}
	\label{fig_ImageGen_1BNtransfer}
\end{figure}

\if false
\begin{figure}[!t]
\setlength{\abovecaptionskip}{0pt}
\setlength{\belowcaptionskip}{-2pt}
\setlength{\abovedisplayskip}{-2pt}
\setlength{\belowdisplayskip}{-2pt}
	\centering
	\includegraphics[width=0.49\textwidth]{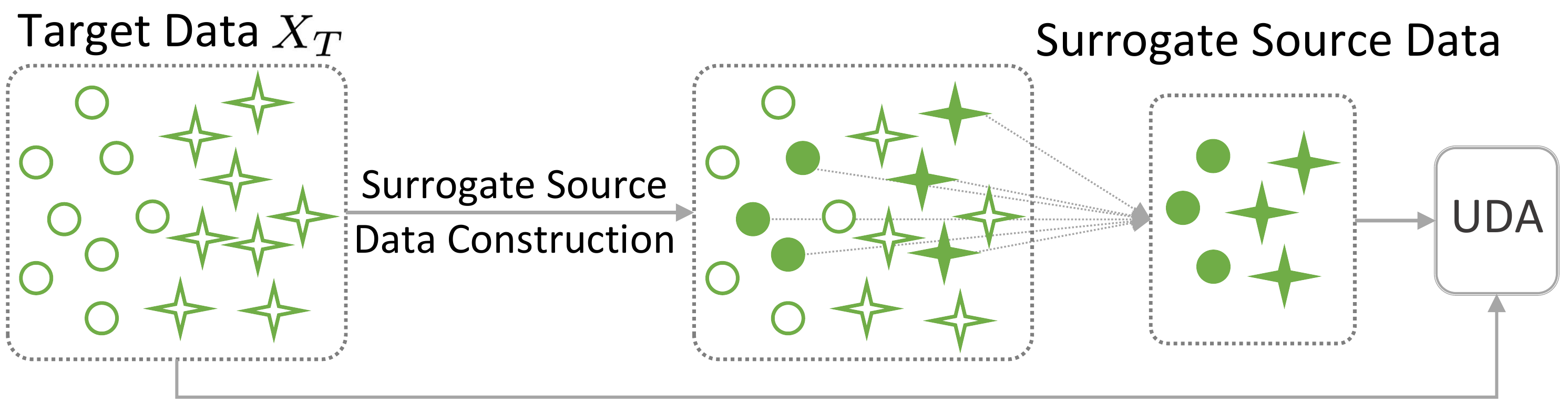}
	\caption{The illustration of \emph{Surrogate Source Data Construction} methods for source data generation.
	The related works construct surrogate/proxy source data by selecting appropriate samples from the target domain directly, and then they perform standard unsupervised domain adaptation.}
	\label{fig_ImageGen_2Surro}
\end{figure}
\fi

(\textbf{2}) \textbf{Surrogate Source Data Construction}. 
To compensate for the inaccessible source domain, some studies~\cite{tian2021source, ding2022proxymix, ye2021source, du2021generation, yao2021sourcesurro} \emph{construct surrogate/proxy source data by selecting appropriate samples from the target domain} directly, as illustrated in Fig.~\ref{fig_ImageGen_2Surro}. 
For example, Tian~\etal~\cite{tian2021source} construct pseudo source samples directly from the provided target samples under the guidance of a designed \emph{sample transport rule}.
The adaptation step and sample transport learning step are performed alternately to refine the approximated source domain and attain confident labels for target data, thus achieving effective cross-domain knowledge adaptation.
Ding~\etal~\cite{ding2022proxymix} build a category-balanced surrogate source domain using pseudo-labeled target samples based on a \emph{prototype similarity measurement}.
During model adaptation, intra-domain and inter-domain mixup regularizations are introduced to transfer label information from the proxy source domain to the target domain, as well as simultaneously eliminate negative effects caused by noisy labels.
Ye~\etal~\cite{ye2021source} select target samples with \emph{high prediction confidence} to construct a virtual source set that mimics source distribution.
To align the target and virtual domains, they develop a weighted adversarial loss based on distribution and an uncertainty measurement to achieve cross-domain adaptation.
Moreover, an uncertainty-aware self-training mechanism is proposed to iteratively produce the pseudo-labeled target set to further enhance adaptation performance.
Du~\etal~\cite{du2021generation} construct a surrogate source domain by first selecting target samples \emph{near the source prototypes} based on an \emph{entropy criterion}, and then enlarging them by a mixup augmentation strategy~\cite{zhang2017mixup}.
The adversarial training is then used to explicitly mitigate cross-domain distribution gap.
Yao~\etal~\cite{yao2021sourcesurro} simulate proxy source domain by freezing the source model and minimizing a supervised objective function for optimization.
For the simulated source set, global fitting is enforced by a model gradient based equality constraint, which is optimized by an alternating direction method of multipliers algorithm~\cite{boyd2011distributed}.

\begin{figure}[!t]
\setlength{\abovecaptionskip}{0pt}
\setlength{\belowcaptionskip}{-2pt}
\setlength{\abovedisplayskip}{-2pt}
\setlength{\belowdisplayskip}{-2pt}
	\centering
	\includegraphics[width=0.49\textwidth]{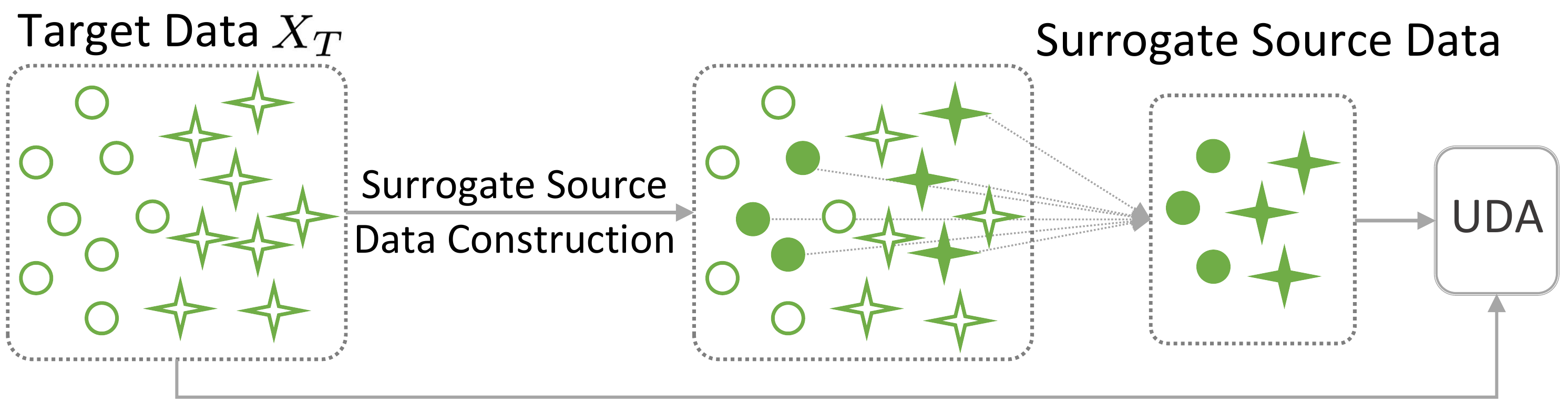}
	\caption{Illustration of \emph{Surrogate Source Data Construction} methods for source data generation.
	These methods first construct surrogate/proxy source data by selecting appropriate samples from the target domain and then perform standard unsupervised domain adaptation (UDA).}
	\label{fig_ImageGen_2Surro}
\end{figure}

\if false
\begin{figure}[!t]
\setlength{\abovecaptionskip}{0pt}
\setlength{\belowcaptionskip}{-2pt}
\setlength{\abovedisplayskip}{-2pt}
\setlength{\belowdisplayskip}{-2pt}
	\centering
	\includegraphics[width=0.49\textwidth]{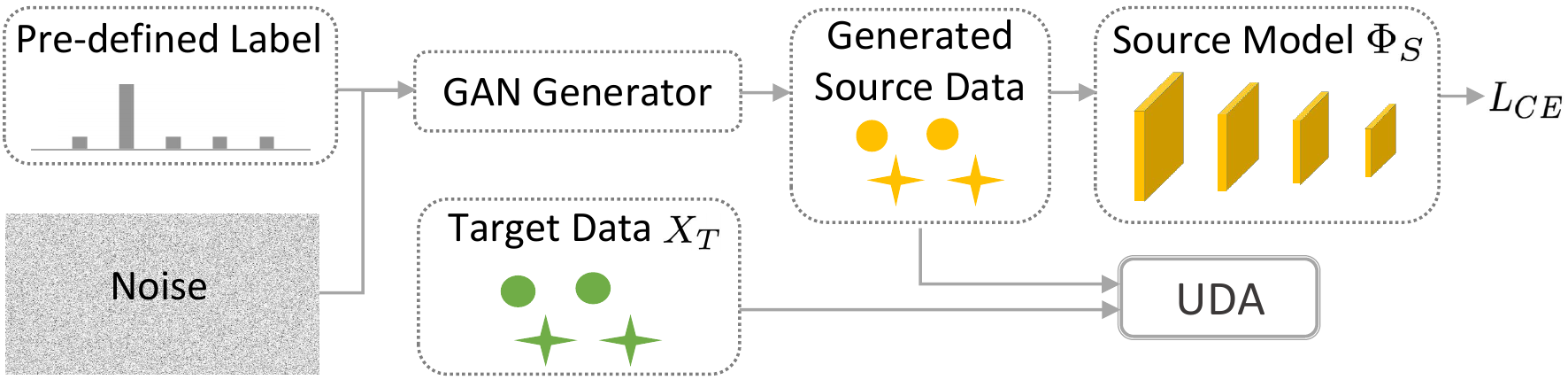}
	\caption{
	The illustration of \emph{GAN-based Image Generation} methods for source data generation.
	Typically, a pre-defined label and random noise acts as the inputs of a GAN-based generator.
	By utilizing the pre-trained source model, they synthesize the source data for cross-domain adaptation.
	$L_{CE}$: cross-entropy loss function.}
	\label{fig_ImageGen_3GANgen}
\end{figure}
\fi

(\textbf{3}) \textbf{GAN-based Image Generation}. 
Instead of approximating the source domain directly using existing target data, Kurmi~\etal~\cite{kurmi2021domain} simulate the source data by training a GAN-based generator, as illustrated in Fig.~\ref{fig_ImageGen_3GANgen}.
Specifically, they first use a \emph{parametric conditional GAN} to generate labeled proxy source data by treating the source classifier as an energy based function.
Then, they learn feature patterns that are invariant across two domains via standard adversarial learning for further adaptation.
Hou~\etal~\cite{hou2021visualizing} also update an \emph{image generator framework} but they aim to translate target images into the source-style ones instead of using the latent noise as in~\cite{kurmi2021domain}.
In their method, the knowledge adaptation is achieved by training 1) a knowledge distillation loss that mitigates the difference between features of newly generated source-style images and those of target images, and 2) a relation-preserving loss that maintains channel-level relationship across different domains. 
Li~\etal~\cite{li2020model} propose \emph{a GAN-embedded generator} conditioned on a pre-defined label to generate target-style data.
By incorporating real target samples, the learnable parameters of the generator and the adapted model can be updated in a collaborative manner.
Moreover, two constraints, \ie, weight regularization and clustering-based regularization, are utilized during model adaptation to preserve source knowledge and ensure high-confident target prediction, respectively.

\subsubsection{Domain Distribution Generation}\label{sec_domain_distribution_generation}
Instead of generating source-like images directly, some studies propose to align feature prototypes or feature distribution of source data~\cite{qiu2021source, tian2021vdm, ding2022source, stan2021unsupervised, stan2021privacy} with those in the target domain.
Specifically, Qiu~\etal~\cite{qiu2021source} \emph{generate feature prototypes} for each source category based on a conditional generator and produce pseudo-labels for the target data. 
The cross-domain prototype adaptation is achieved by aligning the features derived from pseudo-labeled target samples to source prototype with the same category label via contrastive learning.
Tian~\etal~\cite{tian2021vdm} construct a virtual domain by simply sampling from an \emph{approximated gaussian mixture model} (GMM) to mimic unseen source domain distribution.
In terms of adaptation procedure, they reduce the distribution gap between the constructed virtual domain and the target domain via adversarial training, thus bypassing inaccessible source domain.
Their practice is based on the assumption that the feature prototype of each category can be mined from each row of the source classifier' weights~\cite{chen2019closer}.
With the same assumption, Ding~\etal~\cite{ding2022source} leverage such source classifier weights and reliable target pseudo-labels derived by spherical k-means clustering to estimate source feature distribution.
After that, proxy source data can be sampled from the estimated source distribution, and a conventional domain adaptation strategy~\cite{kang2019contrastive} is used to explicitly perform cross-domain feature distribution alignment.
Stan~\etal~\cite{stan2021unsupervised, stan2021privacy} propose to first generate a prototypical distribution representing the source data in an embedding feature space via GMM, and then perform source-free adaptation by enforcing distribution alignment between source and target domains via sliced Wasserstein distance~\cite{lee2019sliced}.

\begin{figure}[!t]
\setlength{\abovecaptionskip}{0pt}
\setlength{\belowcaptionskip}{-2pt}
\setlength{\abovedisplayskip}{-2pt}
\setlength{\belowdisplayskip}{-2pt}
	\centering
	\includegraphics[width=0.49\textwidth]{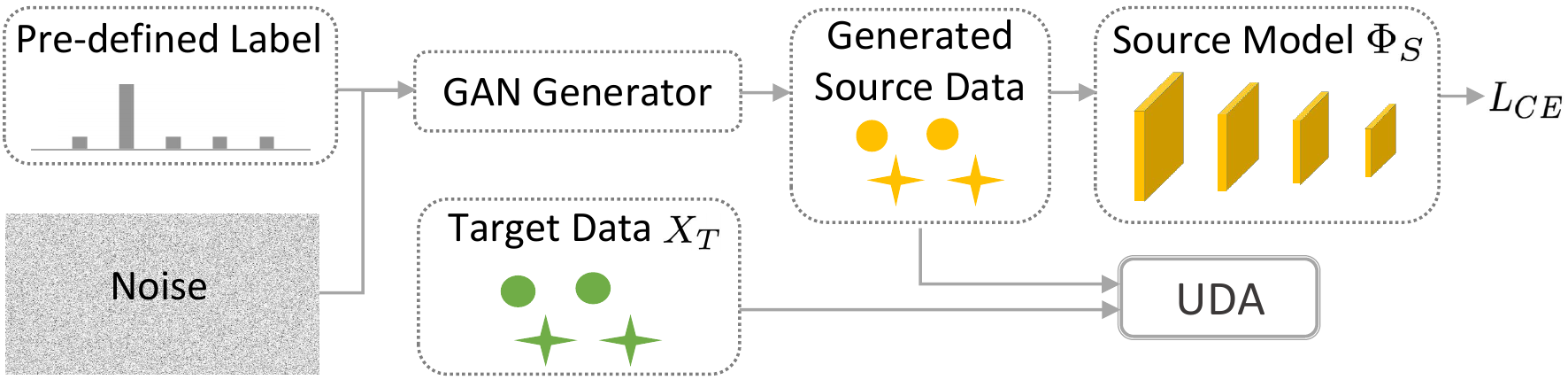}
	\caption{
	Illustration of \emph{Generative Adversarial
Network (GAN) based Image Generation} methods for source data generation.
	Typically, a pre-defined label and random noise act as the inputs of a GAN-based generator.
	By utilizing the pre-trained source model, they synthesize source data for cross-domain adaptation.
	$L_{CE}$: Cross-entropy loss function.}
	\label{fig_ImageGen_3GANgen}
\end{figure}

\subsubsection{Challenges and Insight}
We classify existing domain image generation methods for SFUDA into three subcategories. 
We present the challenges of methods in each subcatetory and our insights below. 
\begin{itemize}[leftmargin=*]
\item Among the above-mentioned three subcategories, the first one (\ie, \emph{batch normalization statistics transfer}) explicitly performs BN statistics matching between source and target domains for style transfer.
Since the BN statistics of the source model are off-the-shelf, these methods are generally efficient and don't require complex model training.
However, BN statistics mainly focus on keeping the style features while the content information cannot be well preserved.
Therefore, this strategy is more applicable to scenarios where the contextual structure of images between source and target domains does not differ too much. 
It may not show good adaptation performance, \eg, from a natural image to a cartoon image, since the content information has significant changes.
Note that BN statistics transfer can also be used as a pre-processing step in source-free domain adaptation, and it can be combined with other strategies, \eg, circular learning~\cite{hong2022source}, for more effective knowledge transfer.

\item Methods in the second subcategory (\ie, \emph{surrogate source data construction}) aim to approximate the proxy source domain using appropriate target samples directly, followed by conventional unsupervised domain adaptation.
Their application is quite broad, including semantic segmentation~\cite{ye2021source}, object recognition~\cite{ding2022proxymix, du2021generation, yao2021sourcesurro}, image classification~\cite{tian2021source}, and digital recognition~\cite{tian2021source, du2021generation}.
In general, methods in this group are straightforward and computation-efficient by avoiding introducing extra hyperparameters, which is different from generative models.
However, because the proxy source samples are directly selected from the target domain, these generated source data may not
effectively represent the original source domain.
Moreover, how to effectively select informative target data for source data approximation is an important topic to be investigated.
Some studies have proposed various strategies for target data selection 
based on entropy measurement~\cite{ye2021source}, source prototype~\cite{ding2022proxymix, du2021generation}, aggregated source decision boundary~\cite{tian2021source}, and equality constrained optimization~\cite{yao2021sourcesurro}.
This is still an open but very interesting future direction. 
For multi-source settings, it is promising to study which source predictor(s) we should refer to for effective target data selection.

\item Methods in the third category (\ie, \emph{GAN-based image generation}) typically synthesize images based on a generative model.
Since the generator can model underlying complex distribution of source data with given random noise, GAN-based models generally create more diverse images compared with methods in second category (\ie, surrogate source data construction).
However, these methods introduce additional frameworks and learnable parameters (\eg, generators and discriminators), which may cost more computation resources.
By comparing experimental results, we find the surrogate source data construction methods~\cite{du2021generation, yao2021sourcesurro} generally outperform the GAN-based generators~\cite{kurmi2021domain, li2020model}.
The possible reason may be that the constructed source data in the former are closer to real data distributions, while those recovered in GAN-based methods usually suffer from a mode collapse problem~\cite{ding2022proxymix} that leads to low-quality images. 
Note that the mode collapse problem can be partly mitigated by using a carefully tuned learning rate, manifold-guided training~\cite{bang2021mggan}, and virtual mapping~\cite{abusitta2021virtualgan}, which is worth exploring further. 
\end{itemize}

Different from image generation methods (Section~\ref{sec_domain_image_generation}) that directly generate source/target-like images, the distribution generation methods (Section~\ref{sec_domain_distribution_generation}) generate feature prototype/distribution to achieve cross-domain feature alignment.
By comparing the reported experimental results, we find that the distribution generation approaches~\cite{qiu2021source, tian2021vdm, ding2022source} usually outperform the GAN-based image generation method~\cite{li2020model}.
And surrogate source data construction methods
~\cite{ding2022proxymix, du2021generation} usually show superior performance compared with the distribution generation methods~\cite{qiu2021source, tian2021vdm}. 
The underlying reason could be that the source distributions directly derived from the existing target data~\cite{ding2022proxymix, du2021generation} are more accurate and stable than the approximated ones~\cite{qiu2021source, tian2021vdm}.
How to drive the approximated source distribution to the real one can be further explored in the future.

\subsection{Model Fine-Tuning Method}\label{sec_model_finetune}
Instead of generating source-like data for standard unsupervised domain adaptation, many studies attempt to fine-tune a pre-trained source model by exploiting unlabeled target data in a self-supervised training scheme.
Based on different strategies for fine-tuning the source model, we divide existing studies into five subcategories: (1) self-supervised knowledge distillation, (2) domain alignment via statistics, (3) contrastive learning, (4) uncertainty-guided adaptation, and (5) hidden structure mining methods, as shown in Fig.~\ref{fig_category_framework}. 
More details are introduced in the following. 

\subsubsection{Self-Supervised Knowledge Distillation}\label{sec_white_sskd}

\begin{figure}[!t]
\setlength{\abovecaptionskip}{0pt}
\setlength{\belowcaptionskip}{-2pt}
\setlength{\abovedisplayskip}{-2pt}
\setlength{\belowdisplayskip}{-2pt}
	\centering
	\includegraphics[width=0.49\textwidth]{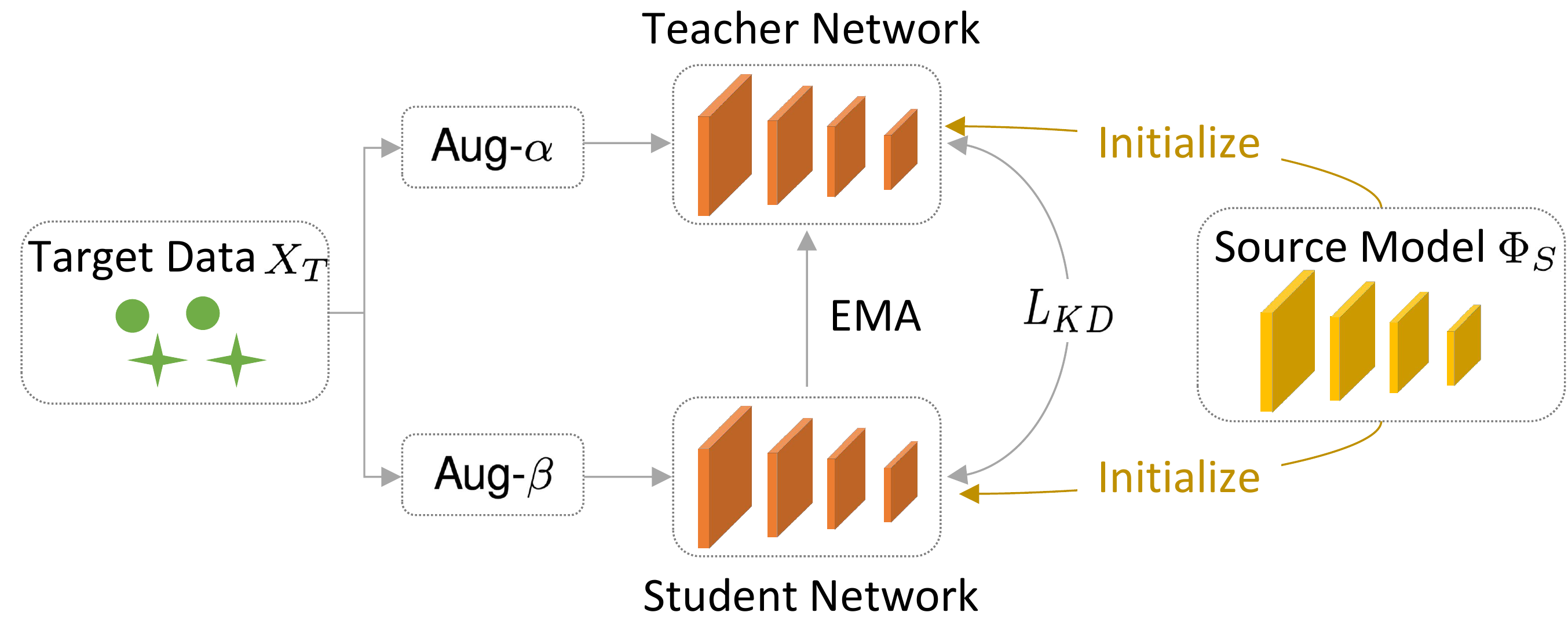}
	\caption{Illustration of \emph{Self-Supervised Knowledge Distillation} methods for source-free unsupervised domain adaptation.
	With target data from different augmentations as inputs, a teacher-student framework is utilized to exploit target features, where parameters of teacher network are usually exponential moving average (EMA) of those of student network.
	Aug-$\alpha$ and Aug-$\beta$ denote two data augmentation methods (\eg, flip, rotation, shift, noise addition, distortion, etc), respectively. 
	$L_{KD}$: Knowledge distillation loss function.
	}
	\label{fig_ModelFinetune_1SSKD}
\end{figure}

Many studies~\cite{liu2022sourcepolyp, yang2021transformer, xiong2021source, chen2021self, liu2021graph, yu2022source, tang2021model, vs2022target} transfer knowledge learned from source data to the target model via knowledge distillation in a self-supervised manner, as illustrated in Fig.~\ref{fig_ModelFinetune_1SSKD}. 
In these works, most of them~\cite{liu2022sourcepolyp, yang2021transformer, xiong2021source, chen2021self, liu2021graph} achieve source-free domain adaptation via \emph{a mean-teacher scheme} for knowledge transfer~\cite{tarvainen2017mean}, where the target model not only learns from unseen target domain but also well preserves source model information.
For instance, Liu~\etal~\cite{liu2022sourcepolyp} propose a self-supervised distillation scheme for automatic polyp detection.
By means of keeping output consistency of weak and strong augmented polyp images, source knowledge is implicitly transferred to the target model with a mean teacher strategy~\cite{tarvainen2017mean}.
Besides, a diversification flow paradigm is designed to gradually eliminate the style sensitivity among different domains, further enhancing model robustness towards style diversification.
Yang~\etal~\cite{yang2021transformer} also propose a self-supervised mean-teacher approach for knowledge distillation, with a Transformer~\cite{dosovitskiy2020image} embedded.
This helps the target model focus on object regions rather than less informative background in an image, thus improving model generalizability.
Assuming that both source and target images are generated from a domain-invariant space by adding noise perturbations on each specific domain, Xiong~\etal~\cite{xiong2021source} establish \emph{a super target domain} via augmenting perturbations based on original target domain.
The super and original target domains are fed into a mean-teacher framework, with three consistency regularization terms (w.r.t. image, instance, and class-wise) introduced for domain alignment.
Chen~\etal~\cite{chen2021self} first divide the target data into clean and noisy subsets guided by a computation loss and regard them as labeled and unlabeled examples, and then utilize the mean teacher technique to self-generate pseudo-labels for the unlabeled target data for domain adaptation.

Instead of utilizing the conventional one-teacher one-student paradigm, Liu~\etal~\cite{liu2021graph} construct a \emph{multi-teacher multi-student} framework, where each teacher/student network is initialized using a public network pre-trained on a single dataset.
Here, a graph is constructed to model the similarity 
among samples, and such relationship predicted by the teacher networks is used to supervise the student networks via a mean-teacher technique.
Rather than leverage the mean-teacher paradigm that averages student's weights, Yu~\etal~\cite{yu2022source} propose to distill knowledge from teacher to student networks by style and structure regularizations, as well as physical prior constraints.
Instead of employing a teacher-student network as the studies mentioned above, Tang~\etal~\cite{tang2021model} achieve data-free adaptation through \emph{gradual knowledge distillation}.
Specifically, they first generate pseudo-labels via a constructed neighborhood geometry, and then use pseudo-labels obtained from the latest epoch to supervise the current training epoch for knowledge transfer.

\begin{figure}[!t]
\setlength{\abovecaptionskip}{0pt}
\setlength{\belowcaptionskip}{-2pt}
\setlength{\abovedisplayskip}{-2pt}
\setlength{\belowdisplayskip}{-2pt}
	\centering
	\includegraphics[width=0.49\textwidth]{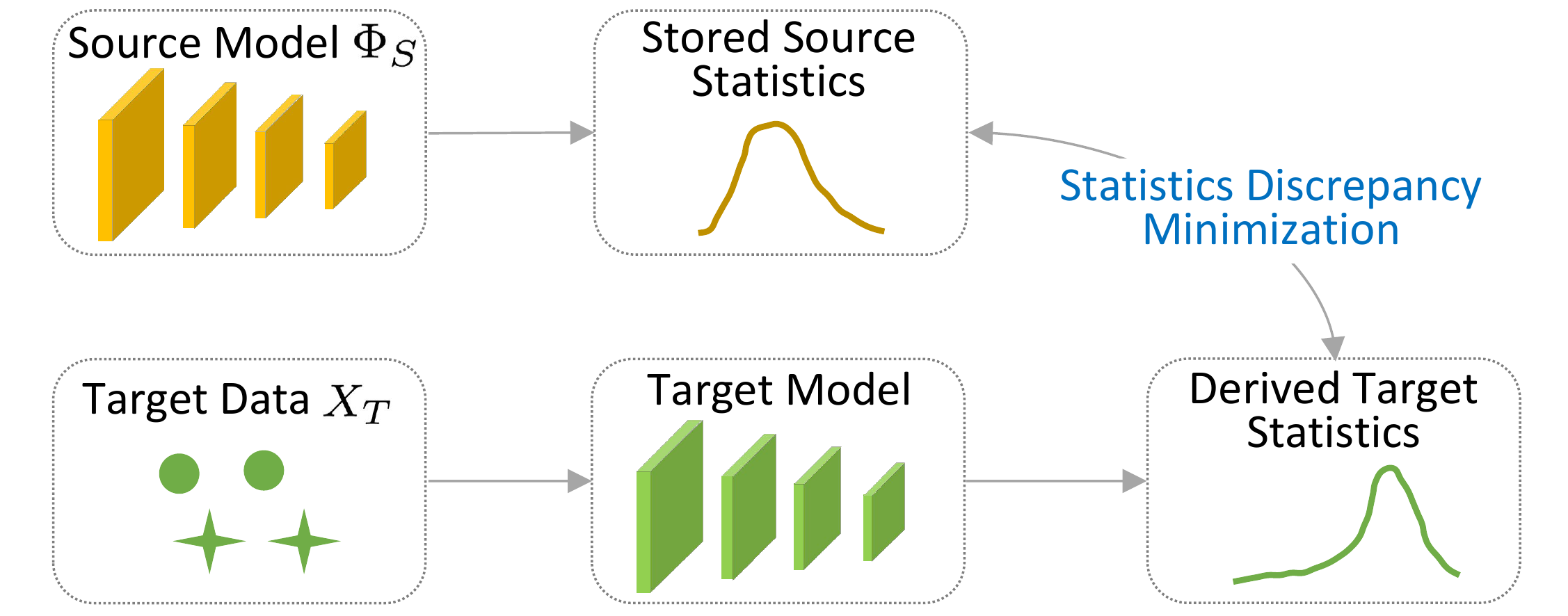}
	\caption{Illustration of \emph{Domain Alignment via Statistics} methods for source-free unsupervised domain adaptation.
	The corresponding methods leverage batch statistics stored in the pre-trained source model to approximate the distribution of inaccessible source data, and then perform cross-domain adaptation by reducing distribution discrepancy between source and target domains.}
	\label{fig_ModelFinetune_2statis}
\end{figure}
\subsubsection{Domain Alignment via Statistics}
Many studies~\cite{ishii2021source, liu2021adapting, paul2021unsupervised, fan2022unsupervised, eastwood2021source, zhang2021source, klingner2022unsupervised} leverage batch statistics stored in the pre-trained source model to approximate the distribution of inaccessible source data, and then perform cross-domain adaptation by reducing distribution discrepancy between source and target domains, as demonstrated in Fig.~\ref{fig_ModelFinetune_2statis}. 
For example, Ishii~\etal~\cite{ishii2021source} approximate feature distribution of inaccessible source data by using \emph{batch normalization statistics (mean and variance)} saved in the pre-trained source model.
Then, Kullback-Leibler (KL) divergence is utilized to minimize the distributional discrepancy between source and target domains, thus achieving domain-level alignment.
Inspired by~\cite{chang2019domain, li2016revisiting}, Paul~\etal~\cite{paul2021unsupervised} update the mean and variance of BatchNorm~\cite{ioffe2015batch} or InstanceNorm~\cite{ulyanov2016instance} of the pre-trained model based on unseen target data. 
Not limited to matching low-order batch-wise statistics (\eg, mean and variance), Liu~\etal~\cite{liu2021adapting} additionally incorporate \emph{high-order batch-wise statistics}, such as scale and shift parameters, to explicitly keep cross-domain consistency.
Moreover, they quantify each channel's transferability based on its inter-domain divergence and assume that the channels with lower divergence contribute more to domain adaptation.
Fan~\etal~\cite{fan2022unsupervised} propose to \emph{align domain statistics adaptively} by modulating a learnable blending factor.
By minimizing the total objective function, each BN layer can dynamically obtain its own optimal factor, which controls the contribution of each domain to BN statistics estimation.
The methods mentioned above are all based on Gaussian-based statistics domain alignment, while Eastwood~\etal~\cite{eastwood2021source} attempt to align \emph{histogram-based statistics} of the marginal feature distributions of the target domain with those stored in the pre-trained source model, thus well extending adaptation to non-Gaussian distribution scenarios.

\subsubsection{Contrastive Learning}
\begin{figure}[!t]
\setlength{\abovecaptionskip}{0pt}
\setlength{\belowcaptionskip}{-2pt}
\setlength{\abovedisplayskip}{-2pt}
\setlength{\belowdisplayskip}{-2pt}
	\centering
	\includegraphics[width=0.45\textwidth]{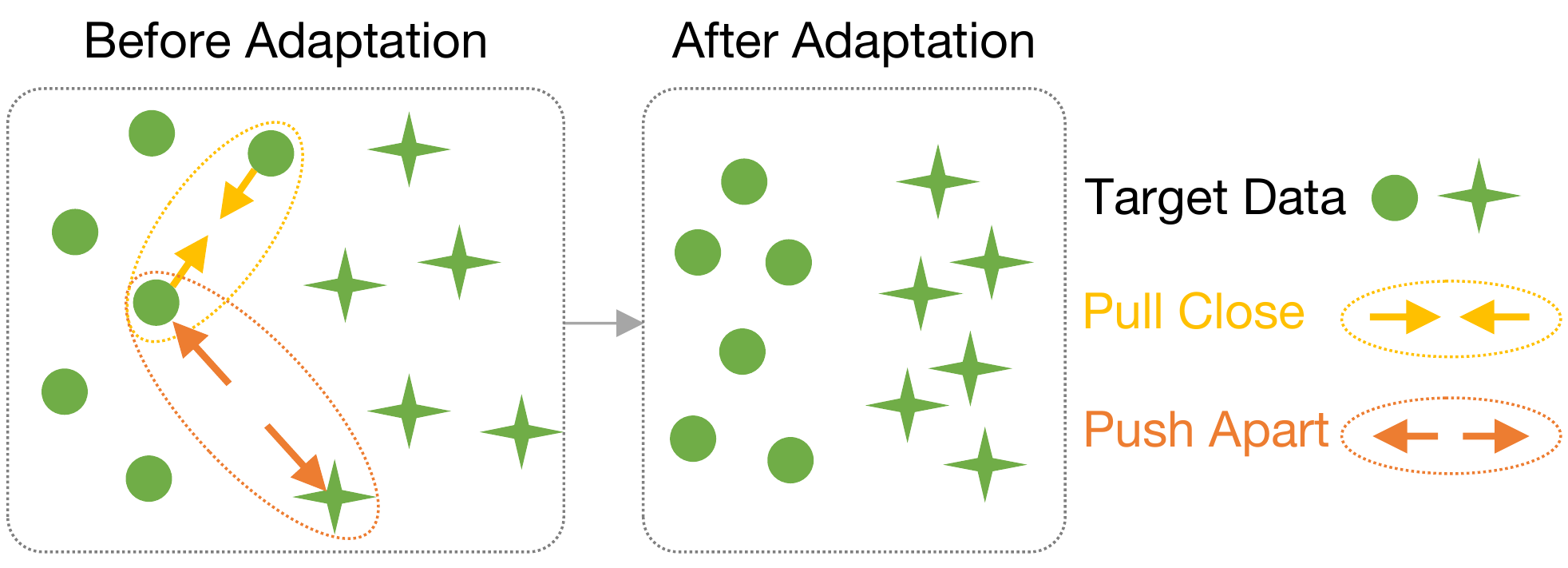}
	\caption{Illustration of \emph{Contrastive Learning} methods for source-free unsupervised domain adaptation.
	These methods exploit discriminative representations among unlabeled target data by pulling instances of similar categories closer and
pushing instances of different categories away 
in feature space.}
	\label{fig_ModelFinetune_3constrast}
\end{figure}

Many contrastive learning studies~\cite{xia2021adaptive, wang2022cross, huang2021model,agarwal2022unsupervised, zhao2022adaptive, liu2022sourceCon} perform data-free adaptation, 
which helps the target model capture discriminative representations among unlabeled target data.
The main idea is to \emph{pull instances of similar categories  closer} and \emph{push instances of different categories away} in feature space, as illustrated in Fig.~\ref{fig_ModelFinetune_3constrast}.

For instance, Xia~\etal~\cite{xia2021adaptive} first adaptively divide target instances into source-similar and source-dissimilar sets, and then design a \emph{class-aware contrastive module} for cross-set distribution alignment.
The idea is to enforce the compactness of target instances from the same category and reduce cross-domain discrepancy, thus prompting effective knowledge transfer from the source model to target data. 
Wang~\etal~\cite{wang2022cross} present a cross-domain contrastive learning paradigm, which aims to minimize the distance between an anchor instance from one domain and instances 
from other domains that share the same category as the anchor.
Due to the unavailability of source data, they utilize source prototypical representations, \ie,  weight vectors in
the classifier layer of a pre-trained 
source model, for feature alignment across two domains.
Huang~\etal~\cite{huang2021model} tackle the data-free domain adaptation by taking advantage of the \emph{historical source hypothesis}.
Specifically, they propose a historical contrastive instance discrimination strategy to learn from target samples by contrasting their learned embeddings generated by the currently adapted and historical models. 
And they also design a historical contrastive category discrimination strategy to weight pseudo-labels of target data to learn category-discriminative target representations, by calculating the \emph{consistency between the current and historical model predictions}.
The two discrimination strategies help exploit historical source knowledge, bypassing the dependence on source data.
Inspired by~\cite{sohn2016improved}, Agarwal~\etal~\cite{agarwal2022unsupervised} introduce a pair-wise contrastive objective function to reduce intra-category distance and meanwhile increase inter-category distance based on generated target pseudo-labels. 
They also introduce robust source and target models by taking advantage of the generated adversarial instances, which facilitates robust transfer of source knowledge to the target domain.

\subsubsection{Uncertainty-Guided Adaptation}
\begin{figure}[!t]
\setlength{\abovecaptionskip}{0pt}
\setlength{\belowcaptionskip}{-2pt}
\setlength{\abovedisplayskip}{-2pt}
\setlength{\belowdisplayskip}{-2pt}
	\centering
	\includegraphics[width=0.49\textwidth]{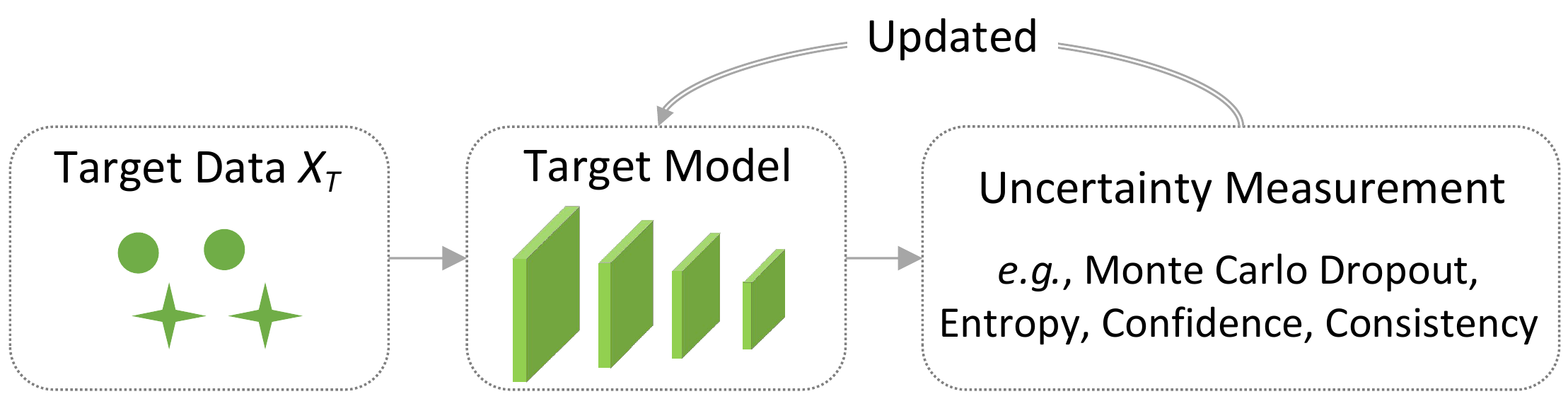}
	\caption{Illustration of \emph{Uncertainty-Guided Adaptation} methods for source-free unsupervised domain adaptation.
	These studies utilize uncertainty to guide target predictions, and such valuable information can be measured by Monte Carlo Dropout, entropy, etc.
	}
	\label{fig_ModelFinetune_4uncertainty}
\end{figure}

Uncertainty can measure how well the target model fits the data distribution~\cite{gawlikowski2021survey}, and many studies~\cite{fleuret2021uncertainty, lee2022feature, chen2021source, xu2022denoising, hegde2021uncertainty, roy2022uncertainty, pei2022uncertainty, li2021free} utilize such valuable information to guide target predictions in source-free adaptation scenarios (see Fig.~\ref{fig_ModelFinetune_4uncertainty}).

For instance, Fleuret~\etal~\cite{fleuret2021uncertainty} estimate uncertainty based on differences between predicted outputs \emph{with and without Dropout operation}~\cite{srivastava2014dropout}.
By minimizing such differences, the prediction uncertainty on target data is reduced, meanwhile the learnable feature abstractor can be more robust to noise perturbations.
Lee~\etal~\cite{lee2022feature} exploit aleatoric uncertainty by \emph{encouraging intra-domain consistency} between target images and their augmented ones and enforcing \emph{inter-domain feature distribution consistency}.
Chen~\etal~\cite{chen2021source} introduce a prediction denoising approach for a cross-domain segmentation task.
In this study, a key component is introducing \emph{pixel-wise denoising via uncertainty evaluation using Monte Carlo Dropout}~\cite{gal2016dropout, kendall2017uncertainties}, which calculates the standard deviation of several stochastic outputs and keeps it under a manually-designed threshold.
In this way, the noisy pseudo-labels can be filtered out, helping improve pseudo-label quality to achieve effective adaptation.
Xu~\etal~\cite{xu2022denoising} also propose an uncertainty-guided pseudo-labeling denoising scheme, but they use soft label correction instead of manually discarding unreliable data points. 
Specifically, they first identify mislabeled data points by utilizing a joint distribution matrix~\cite{xu2021noisy, northcutt2021confident}, and then assign larger confident weights to those with higher certainty based on Monte Carlo Dropout.
Combining target data and the corresponding rectified pseudo-labeling, a commonly used cross-entropy objective function can be leveraged for training the target model.
Sharing the similar idea, Hegde~\etal~\cite{hegde2021uncertainty} allocate lower weights for uncertain pseudo-labels, where the uncertainty is measured by prediction variance based on Monte Carlo Dropout~\cite{gal2016dropout, kendall2017uncertainties}.
Considering that using Monte Carlo Dropout~\cite{gal2016dropout} for uncertainty estimation requires manual hyperparameter adjustment~\cite{gal2017concrete}, Roy~\etal~\cite{roy2022uncertainty} quantify source model's uncertainty using a \emph{Laplace approximation}~\cite{tierney1986accurate, mackay2003information}.
For model training, they assign smaller weights to those target samples that are farther away from source hypothesis (measured by uncertainty), avoiding misalignment of dissimilar samples.
Pei~\etal~\cite{pei2022uncertainty} tackle the uncertainty issue from the perspective of improving \emph{source model transferability}.
Specifically, they estimate channel-aware transferability of the source model to target data based on an uncertainty distance, which measures the closeness between target instances and source distribution.
With the aim of dynamically exploiting the source model and target data, the target model obtains the source knowledge from the transferable channels and neglects those less-transferable ones.
Unlike previous studies, Li~\etal~\cite{li2021free} quantify uncertainty using self-entropy and propose a \emph{self-entropy descent mechanism} to seek the optimal confidence threshold for robust pseudo-labeling of target data.
They also leverage false negative mining and mosaic augmentation~\cite{bochkovskiy2020yolov4} 
to further eliminate the negative influence of noisy labels to enhance adaptation performance.

\begin{figure}[!t]
\setlength{\abovecaptionskip}{0pt}
\setlength{\belowcaptionskip}{-2pt}
\setlength{\abovedisplayskip}{-2pt}
\setlength{\belowdisplayskip}{-2pt}
	\centering
	\includegraphics[width=0.46\textwidth]{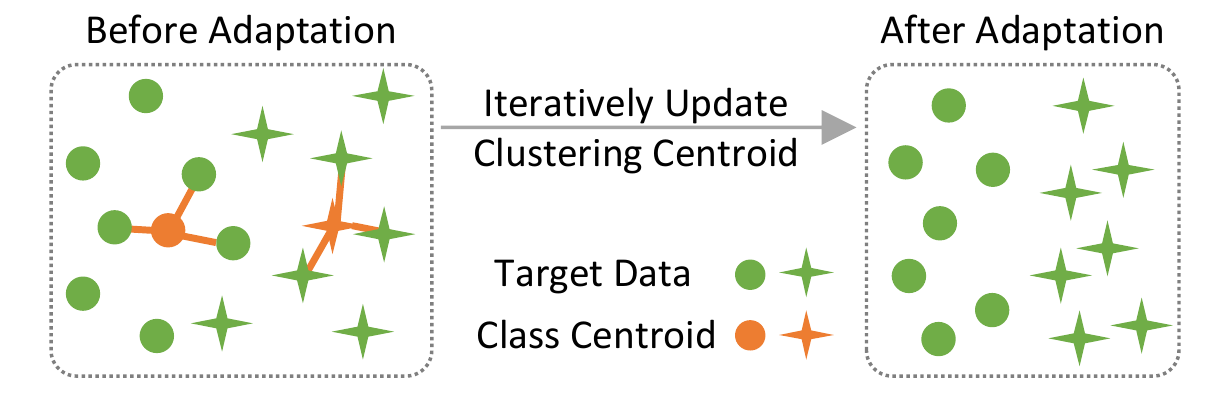}
	\caption{Illustration of \emph{Hidden Structure Mining} methods for source-free unsupervised domain adaptation.
	These methods take into consideration intrinsic feature structures of target domain and iterate between target model refinement and clustering centroid update.}
	\label{fig_ModelFinetune_5structure}
\end{figure}

\subsubsection{Hidden Structure Mining}
Many studies~\cite{yang2021exploiting, yang2021generalized, tang2021nearest, liang2020we, liang2021source, lee2022confidence, li2022jacobian, yang2022attracting} take into consideration intrinsic feature structures of target domain and update the target model via clustering-aware pseudo-labeling. 
In Fig.~\ref{fig_ModelFinetune_5structure}, we illustrate the main idea of hidden structure mining methods. 

For example, Yang~\etal~\cite{yang2021exploiting} observe that target data can intrinsically form a \emph{certain cluster structure} that can be used for 
domain adaptation.
Specifically, they estimate affinity among target data by taking into account the neighborhood patterns captured from local-, reciprocal-, and expanded-neighbors.
Source-free adaptation is achieved by encouraging consistent predictions for those with high affinity.
Similarly, Yang~\etal~\cite{yang2021generalized} also exploit \emph{neighborhood structure information} of target data.
They propose a local structure clustering strategy to encourage prediction consistency among $k$-nearest target features, thus pushing target data with semantically similar neighbors closer.
Tang~\etal~\cite{tang2021nearest} leverage \emph{semantic constraints} hidden in \emph{geometric structure} among target data to encourage robust clustering based on a cognition mechanism~\cite{ashby2005human}.
Source hypothesis transfer (SHOT)~\cite{liang2020we} and SHOT++~\cite{liang2021source} attempt to mine the feature structure of the target domain, but they cannot fully exploit the meaningful context since the used self-supervised pseudo-labeling does not take into account each dimension's covariance in the feature space.
To address this issue, Lee~\etal~\cite{lee2022confidence} utilize GMM in the target domain to obtain data structure, and design a \emph{joint model-data structure score} to concurrently exploit source and target knowledge.
Li~\etal~\cite{li2022jacobian} utilize neighbor structure information from a new aspect by proposing a generic and model smoothness-assisted \emph{Jacobian norm regularization} term, which is used to manipulate the consistency between each target instance and its neighbors.
This Jacobian norm regularizer can be easily plugged into existing source-free domain adaptation frameworks for boosting performance. 

Different from the above mentioned methods, some studies tackle source-free domain adaptation from other perspectives.
Li~\etal~\cite{li2021divergence} achieve data-free adaptation from an \emph{adversarial-attack aspect}, which aims to generate adversarial target instances by adding diverse perturbations to attack the target model.
Then, mutual information maximization is performed between representations extracted by the source and target model for the same target instance.
The above two steps are performed alternatively, by which the domain-invariant source knowledge can be preserved and the rich target patterns can be well explored.
Instead of exploring domain-invariant features for cross-domain knowledge transfer, Wang~\etal~\cite{wang2022exploring} mine \emph{domain-invariant parameters} stored in the source model.
They assume that only partial domain-invariant parameters of the source model contribute to domain adaptation, and their goal is to capture such parameters while penalizing the domain-specific ones.
Liang~\etal~\cite{liang2019distant} explore source-free adaptation from the perspective of \emph{minimum centroid shift}, with the aim of searching a subspace where target prototypes are mildly shifted from source prototypes.
An alternating optimization scheme is leveraged for model convergence and target pseudo-label update.
Inspired by maximum classifier discrepancy~\cite{saito2018maximum}, Yang~\etal~\cite{yang2020casting} introduce an \emph{auxiliary bait classifier} for cross-domain feature alignment combined with the source anchor classifier.
These two classifiers aim to collaboratively push uncertain target representations to the correct side of the source classifier boundary.

\subsubsection{Challenges and Insight}
We classify existing model fine-tuning methods for SFUDA into five subcategories. 
The challenges of methods in each subcatetory and our insights are presented below. 

\begin{itemize}[leftmargin=*]
\item
The methods in the first subcategory, \ie, \emph{self-supervised knowledge distillation}, interpret source-free domain adaptation as a knowledge extraction and transfer process, aiming to learn domain-invariant feature representations.
Most exiting studies transfer source knowledge to the target model via a mean teacher strategy~\cite{tarvainen2017mean}, where teacher weights are an exponential moving average of student weights.
Hence, model parameters of both teacher and student networks are tightly coupled, which may lead to a performance bottleneck.
A possible solution is to introduce a dual-student framework and let one student learn features flexibly, which may disentangle teacher-student weights to some extent~\cite{ke2019dualstu}.

\item The second subcategory, \ie, \emph{domain alignment via statistics}, leverages batch statistics stored in a pre-trained source model to approximate distribution of inaccessible source data.
Compared with other categories, these statistics-based methods are lightweight and prone to generalize to other tasks, since they require only a few update steps of batch-wise statistics parameters and are potentially applicable to real-time deployment~\cite{klingner2022unsupervised}.
However, they are not suitable for problems that use deep network architectures without batch normalization layers.

\item The methods in the third subcategory, \ie, \emph{contrastive learning}, aim to bring similar-class samples closer and push dissimilar-class samples apart based on generated target pseudo-labels.
Therefore, if the pseudo-labels contain much noise, these methods may suffer from substantial performance degradation.
Moreover, a memory bank is usually required to store the similarity relationship between current and historical feature representations of target data, which could bring memory burden.
It is interesting to investigate the storage- and transmission-efficient contrastive learning strategies in source-free settings.
In addition, several recent studies~\cite{chen2020simple, he2020momentum} have shown that data pair construction is crucial for effective contrastive learning.
One solution is utilizing contrastive information between target data and their augmented ones.
Previous studies~\cite{chen2020simpleaug} often use either strong or weak transformations for data augmentation, where strong augmentations mostly distort the structures of original images (\eg, shape distortion) while weak augmentations usually limit transformations to preserve the images' structures (\eg, flip).
Here we propose to dynamically mix strong and weak augmentation of target data, which may help learn more robust representations.

\item The methods in fourth subcategory, \ie, \emph{uncertainty-guided adaptation}, focus on reducing prediction uncertainty of target data.
Many studies~\cite{chen2021source, xu2022denoising} use Monte Carlo Dropout for uncertainty estimation, but this technique requires specialized network architecture design and model training, bringing troublesome hyperparameter tuning~\cite{gal2017concrete}.
A recent study~\cite{chen2021source} points out that their method can only handle problems with minor domain shift, and performs poorly on problems with severe domain shift.
It is interesting to explore this challenging problem in the future.

\item The last subcategory, \ie, \emph{hidden structure mining}, considers intrinsic clustering structure of the target domain, assuming that geometric structure of target data may provide informative context~\cite{tang2021nearest}.
The advantage of these methods is that no auxiliary frameworks are required, and thus, they can be easily incorporated into other adaptation frameworks.
However, these methods have at least three disadvantages.
(1) Most existing studies need to iterate between feature clustering and model update, which may hinder training efficiency and cause a memory burden.
(2) These methods may be infeasible to handle extremely large-scale datasets due to the difficulty of saving global latent feature embeddings of the whole dataset~\cite{chen2020unsupervisedre}.
(3) Most studies construct target geometric structures in Euclidean space, which may not be suitable for problems with non-Euclidean data such as graphs. 
Thus, how to improve training efficiency and deal with the large-size dataset, as well as mining geometry information of non-Euclidean data deserve further research. 
\end{itemize}

From the application perspective, computation-efficient approaches are more applicable for pixel-wise semantic segmentation tasks, which require higher resources compared with classification tasks. 
And those memory-intensive approaches such as contrastive learning may be not suitable for semantic segmentations.
Moreover, it is worth noting that data generation methods detailed in Section~\ref{sec_data_generation} can be used in conjunction with the model fine-tuning methods described in this section.
For instance, one can first generate a virtual source domain by selecting appropriate target samples, and thus a standard unsupervised domain adaptation framework could be applied.
To further exploit target information, we then take account of geometric structure of target samples and generate corresponding target pseudo-labels to fine-tune the target model.
These two steps can be optimized iteratively, helping generate more representative source domain and refine the target model.

\section{Black-Box Source-Free Unsupervised Domain Adaptation}\label{sec_black_SFUDA}
Different from white-box methods, \emph{in the setting of black-box source-free domain adaptation, both the source data $\{X_S, Y_S\}$ and detailed parameters of the source model $\Phi_S$ are not accessible}.
Only the hard or soft model predictions of the target data $X_T$ from the source model $\Phi_S$ are leveraged for domain adaptation.
Depending on the utilization of the black-box predictor, the existing black-box SFUDA studies can be mainly divided into three categories: \textbf{Self-Supervised Knowledge Distillation}, \textbf{Pseudo-Label Denoising}, and \textbf{Generative Distribution Alignment} methods, with details introduced below. 

\subsection{Self-Supervised Knowledge Distillation}


Some studies~\cite{liang2022dine, liang2021distill, liu2022unsupervised1, liu2022unsupervised2, xu2022extern, peng2022toward} construct a \emph{teacher-student-style network} architecture with knowledge distillation to translate the source knowledge to the target domain in a self-supervised manner.
For instance, Liang~\etal~\cite{liang2022dine, liang2021distill} enforce output consistency between a source model (\ie, teacher) and a customized target model (\ie, student) via a self-distillation loss.
Specifically, a memory bank is first constructed to store the prediction of each target sample based on the black-box source model.
This source model then acts as a teacher to maintain an exponential moving averaging of source and target prediction following~\cite{laine2016temporal, kim2021self}. 
Additionally, structural regularization on the target domain is further incorporated during adaptation for more effective knowledge distillation.
Similarly, Liu~\etal~\cite{liu2022unsupervised1, liu2022unsupervised2}
employ an \emph{exponential mixup decay} scheme to explicitly keep prediction consistency of source and target domains, thus gradually capturing target-specific feature representations and obtaining the target pseudo-labels.
Xu~\etal~\cite{xu2022extern} extend the teacher-student paradigm from image analysis 
to more challenging video analysis, where not only spatial features but also temporal information are taken into consideration during domain adaptation.
For knowledge distillation, the target model is regarded as a student, which aims to learn similar predictions generated by a teacher (\ie, source) model.
The teacher model is meanwhile updated to maintain an exponential moving averaging prediction.
Instead of distilling knowledge between source and target domains, Peng~\etal~\cite{peng2022toward} transfer knowledge between the target network and its subnetwork in a mutual way, where the subnetwork is a slimmer version generated from the original target network following Yang~\etal~\cite{yang2020mutualnet}.
And target features are extracted by leveraging multi-resolution input images, helping improve the generalization ability of the target network.
Moreover, a novel data augmentation strategy, called frequency MixUp, is proposed to emphasize task-related regions-of-interests while simultaneously reducing background interference. 

\subsection{Pseudo-Label Denoising}
Some studies~\cite{zhang2021unsupervised, luo2021exploiting} tackle domain shift by carefully denoising unreliable target pseudo-labels. 
For example, Zhang~\etal~\cite{zhang2021unsupervised} combat noisy pseudo-labels via \emph{noise rate estimation}, which first preserves more training samples at the start of the training process following~\cite{arpit2017closer} and then gradually filters out the noisy ones based on their loss values as training proceeds.
The pseudo-labels are iteratively refined according to a category-dependent sampling strategy, encouraging the model to capture more diverse representations to improve model generalization ability.
Different from Zhang~\etal~\cite{zhang2021unsupervised} that only select part of reliable target data during model training, Luo~\etal~\cite{luo2021exploiting} take into account all target data and rectify noisy pseudo-labels from a \emph{negative learning} aspect.
Specifically, their approach assigns complementary ground-truth labels for each target sample, helping alleviate error accumulation for noisy prediction.
Moreover, a maximum squares objective function is utilized as confidence regularization to prevent the target model from being trapped in easy sample training.
Yang~\etal~\cite{yang2022divide} incorporate \emph{pseudo-label denoising} and \emph{self-supervised knowledge distillation} into a unified framework.
Specifically, domain knowledge is first distilled from the trained source predictor to warm up the target model by an exponential moving averaging scheme. 
The unlabeled target domain is then split into two subsets (\ie, easy and hard groups) according to their adaptation difficulty~\cite{arazo2019unsupervised}, and the MixMatch strategy~\cite{berthelot2019mixmatch} is leveraged to progressively exploit all target representations.
In this way, the noise accumulation is further suppressed, thereby improving the efficacy of pseudo-label denoising.

\subsection{Generative Distribution Alignment}
Different from the above methods, some studies perform distribution alignment across domains in a generative way.
For instance, Yeh~\etal~\cite{yeh2021sofa} perform domain adaptation by maximizing the lower bound in variational inference. 
Specifically, they construct a \emph{generation path} as well as an \emph{inference path}, where the generation path produces a prior feature distribution derived from predicted category labels, and the inference path approximates a posterior feature distribution based on each target instance.
The latent distribution alignment can be achieved by maximizing the evidence lower vound in variational inference for cross-domain adaptation.
Similarly, Yang~\etal~\cite{yang2021model} also construct the generation and inference paths, but they achieve adaptation via minimizing the upper bound of the prediction error of target data in variational inference.
Zhang~\etal~\cite{zhang2022lightweight} achieve source-free adaptation by first building multiple source models and then generating a \emph{virtual intermediate surrogate domain} to select target samples with minimum inconsistency predicted by the source models.
The knowledge transfer is achieved by feature distribution alignment between the virtual surrogate domain and the target domain based on a joint probability maximum mean discrepancy~\cite{zhang2020discriminative}.

\subsection{Challenges and Insight}
In this section, we classify existing black-box SFUDA methods into three categories based on how they utilize the noisy target predictions. 
The challenges of each category and our insights are presented below. 
\begin{itemize}[leftmargin=*]
\item
The first category, \ie, \emph{self-supervised knowledge distillation}, aims to gradually transfer source knowledge to a customized target model by enforcing output consistency between a teacher (source) and a student (target) network.
This learning strategy has also been used in white-box SFUDA (see Section~\ref{sec_white_sskd}). 
The difference is that model weights of student networks are accessible in white-box SFUDA methods, but not in black-box SFUDA.  
In black-box SFUDA, instead of leveraging any parameter details, the teacher network is only updated by source predictions and historical target predictions.
The two items are typically weighted by a momentum factor, which helps dynamically adjust their contributions.
Self-supervised knowledge distillation has shown promising performance in object recognition~\cite{liang2022dine}, semantic segmentation~\cite{liu2022unsupervised1}, and video action recognition tasks~\cite{xu2022extern}.

\item The methods in the second category, \ie, \emph{pseudo-Label denoising}, tackle black-box SFUDA from the perspective of noisy label rectification.
It has shown that a pseudo-label denoising approach~\cite{zhang2021unsupervised} has inferior performance than the self-supervised knowledge distillation method~\cite{liang2022dine}.
The reason may be that the former~\cite{zhang2021unsupervised} only focuses on noisy prediction itself while neglecting target data structure that is well considered in the latter~\cite{liang2022dine}.
Considering that pseudo-label denoising methods can tackle unbalanced label noise via noise rate estimation, combining pseudo-label denoising with self-supervised knowledge distillation strategies will be a promising future direction, especially in class-imbalance scenarios. 
Moreover, if the black-box predictor only provides one-hot hard predictions instead of probability predictions, the utility of methods in this subcategory will be greatly reduced. 
The reason is that the noise rate cannot be well estimated in practice, \eg, there is nearly no difference between the output of $[0.45, 0.55]$ and that of $[0.05, 0.95]$ because the source predictor produces the same output (\ie, $[0, 1]$).

\item The third category, \ie, \emph{generative distribution alignment}, attempts to perform domain adaptation by minimizing feature distribution discrepancy across the source and target domains.
Since source distribution is inaccessible in black-box models, some generative approaches are utilized to generate such reference distribution for target data to align with, including variational autoencoder~\cite{yeh2021sofa, yang2021model} and surrogate source domain construction~\cite{zhang2022lightweight}.
These methods are more suitbale for recognition/classification tasks, but less suitable for 
semantic segmentation tasks.
For example, generating surrogate feature distribution of an object (\eg, car) is usually easier than that of a semantic scene (\eg, cityscape), since the latter contains different objects and thus the pixel-wise neighborhood relationship is difficult to model in practice.
\end{itemize}

Besides the strategies proposed above, it is also crucial to build a \emph{general and robust black-box source model}, with which the target predictions tend to be more accurate.
To achieve that, one possible solution is augmenting the diversity of source data (\eg, adding some perturbation) before constructing the source model, which may eliminate style discrepancy between two domains, thus improving the generalization ability of the source model.
Another solution is using soft probability labels instead of hard one-hot labels (\eg, [0, 1]) for model training, which prevents the source model from being over-confident and helps enhance its generalizability. 
Compared to white-box methods, there are relatively few black-box SFUDA methods as well as benchmark datasets, which needs to be further explored. 


\begin{table*}[!tbp]
	\setlength{\belowcaptionskip}{-2pt}
	\setlength{\abovecaptionskip}{-2pt}
	\setlength\abovedisplayskip{-2pt}
	\setlength\belowdisplayskip{-2pt}
	\scriptsize
	\renewcommand\arraystretch{1.2}
	\caption{Commonly used datasets for evaluating the performance of source-free unsupervised domain adaptation (SFUDA) approaches.}
	\centering
	\begin{tabular*}{1\textwidth}{@{\extracolsep{\fill}} l|c|c|c|l}
	\toprule
	~Dataset & Domain~\# & Instance~\# & Category~\# & Description \\
 \hline
    \cellcolor{gray!30}\textbf{Digit Recognition} \cellcolor{gray!30}& \cellcolor{gray!30}& \cellcolor{gray!30}& \cellcolor{gray!30}& \cellcolor{gray!30} \\
    ~~Digits-Five~\cite{peng2019momentDA} & 5 & 215,695 & 10 & MNIST~\cite{lecun1998gradient}, SVHN~\cite{netzer2011reading}, USPS~\cite{hull1994database}, MNIST-M~\cite{ganin2015unsupervised}, Synthetic Digits~\cite{ganin2015unsupervised}  \\
    
    \rowcolor{gray!30}\textbf{Semantic Segmentation} & & & &\\
    ~~Segmentation datasets & 4 & 45,766 & - & GTA5~\cite{richter2016playing}, Cityscapes~\cite{cordts2016cityscapes}, SYNTHIA~\cite{ros2016synthia}, NTHU~\cite{chen2017nomore} \\ 
 
    \rowcolor{gray!30}\textbf{Object Recognition} & & && \\
    ~~Office-31~\cite{saenko2010adapting} & 3 & 4,652 & 31 & Amazon, Webcam, DSLR \\
    ~~Office-Home~\cite{venkateswara2017deep} & 4 & 15,500 & 65 & Artistic, Clip Art, Product, Real-World \\
    ~~VisDA~\cite{peng2017visda} & 2 & 280,000 & 12 & Synthetic and real images \\
    ~~Office-Caltech-10~\cite{gong2012geodesic} & 4 & 2,533 & 10 & Amazon, DSLR, Webcam, Caltech10 \\
    ~~ImageCLEF-DA~\cite{caputo2014imageclef} & 4 & 2,400 & 12 & Caltech-256~\cite{griffin2007caltech}, ImageNet ILSVRC2012~\cite{deng2009imagenet}, PASCAL VOC2012~\cite{everingham2010pascal}, Bing~\cite{bergamo2010exploiting} \\
    ~~PACS~\cite{li2017deeperDG} & 4 & 9,991 & 7 & Art painting, Cartoon, Photo, Sketch\\
    ~~DomainNet~\cite{peng2019momentDA} & 6 & 600,000 & 345 & Clipart, Infograph, Painting, Quickdraw, Real, Sketch \\
    ~~MiniDomainNet~\cite{zhou2021domain} & 4 & 140,000 & 126 & Clipart, Painting, Real, Sketch \\
    ~~PointDA-10~\cite{qin2019pointdan} & 3 & 33,067 & 10 & ModelNet~\cite{wu20153dshape}, ShapeNet~\cite{chang2015shapenet}, ScanNet~\cite{dai2017scannet} \\
    
    \rowcolor{gray!30}\textbf{Face Anti-Spoofing} & & & &\\
    ~~Face datasets & 4 & 7,130 & - & Replay-Attack~\cite{chingovska2012effectiveness}, OULU-NPU~\cite{boulkenafet2017oulu}, CASIA-MFSD~\cite{zhang2012face}, MSU-MFSD~\cite{wen2015face} \\  

    \rowcolor{gray!30}\textbf{LiDAR Detection} & & & &\\
    ~~LiDAR datasets & 3 & 158,510 & - & Waymo~\cite{sun2020scalability}, KITTI~\cite{geiger2012we}, nuScenes~\cite{caesar2020nuscenes} \\ 
    
    \rowcolor{gray!30}\textbf{Video Action Recognition} & & & &\\
    ~~UCF-HMDB$_{\text{full}}$~\cite{chen2019temporal} & 2 & 3,209 & 12 &	UCF101~\cite{soomro2012ucf101}, HMDB51~\cite{kuehne2011hmdb} \\
    ~~Sports-DA~\cite{xu2021multivideo} & 3 & 40,718 & 23 & UCF10~\cite{soomro2012ucf101}, Sports-1M~\cite{karpathy2014large}, Kinetics~\cite{kay2017kinetics} \\
    ~~Daily-DA~\cite{xu2021multivideo} & 4 & 18,949 & 8 & ARID~\cite{xu2021arid}, HMDB51~\cite{kuehne2011hmdb}, Moments-in-Time~\cite{monfort2019moments}, Kinetics~\cite{kay2017kinetics} \\

   
    \rowcolor{gray!30}\textbf{Traffic Sign Recognition} & & & &\\
    ~~Sign datasets & 2 & 151,839 & 43 & Syn.Signs~\cite{moiseev2013evaluation}, GTSRB~\cite{stallkamp2011german} \\
    
    \rowcolor{gray!30}\textbf{Image Classification} & & & &\\
    ~~VLCS~\cite{fang2013unbiased} & 4 & 10,729 & 5 & Caltech101~\cite{fei2006oneshot}, LabelMe~\cite{russell2008labelme}, SUN09~\cite{choi2010exploiting}, VOC2007~\cite{everingham2010pascal} \\
    
    \rowcolor{gray!30}\textbf{Medical Data} & & & &\\
    ~~BraTS2018~\cite{menze2014multimodal} & 2 & 285 & - & Cross-disease (high and low grade glioma), cross-modality (T1, T1ce, T2, FLAIR)\\
    ~~MMWHS~\cite{zhuang2016multiscale} & 2 & 40 & - & Cross-modality (magnetic resonance imaging, computed tomography) \\
    ~~Brain skull stripping~\cite{li2022sourceskull} & 3 & 35 & - & NFBS~\cite{eskildsen2012beast}, ADNI~\cite{jack2008alzheimer}, dHCP~\cite{makropoulos2018developing} \\
    ~~Polyp segmentation & 4 & 2,718 & - & CVC-ClincDB~\cite{bernal2015wm}, Abnormal Symptoms~\cite{hoang2019enhancing}, ETIS-Larib~\cite{silva2014toward}, EndoScene~\cite{vazquez2017benchmark} \\ 
    ~~EEG MI Classification~\cite{zhang2022lightweight}  & 4 & 528 & 2/4 & MI2-2~\cite{tangermann2012review}, MI2-4~\cite{tangermann2012review}, MI2015~\cite{faller2012autocalibration}, AlexMI~\cite{jayaram2018moabb} \\
    ~~Prostate segmentation & 2 & 682 & - & NCI-ISBI 2013 Challenge~\cite{bloch2015nci}, PROMISE12 challenge~\cite{litjens2014evaluation} \\ 
    ~~Optic disc\&cup segmentation & 3 & 660 & - & REFUGE~\cite{orlando2020refuge}, RIMONE-r3~\cite{fumero2011rim}, Drishti-GS ~\cite{sivaswamy2015comprehensive} \\ 
    ~~Autism diagnosis & 4 & 411 & 2 & NYU, USM, UCLA, and UM of ABIDE dataset~\cite{di2014autism}\\ 
	\bottomrule
	\end{tabular*}
\label{tab_datasetSFUDA}
\end{table*}

\section{Discussion}\label{sec_discussion}
In this section, we first compare the white-box and black-box SFUDA methods and then summarize several useful strategies to improve model generalizability. 
We also list datasets commonly used in the field in Table~\ref{tab_datasetSFUDA}.

\subsection{Comparison of White-Box and Black-Box SFUDA}
By comparing existing white-box and black-box SFUDA methods, we have the following interesting observations. 
\begin{itemize}[leftmargin=*]
\item Compared with black-box SFUDA that cannot access any source parameters, white-box SFUDA is capable of mining more source knowledge (\eg, batch statistics) that facilitates more effective domain adaptation.

\item White-box SFUDA methods may suffer from data privacy leakage problems~\cite{zhang2021unsupervised}. 
For instance, Yin~\etal~\cite{yin2020dreaming} 
reveal that raw data can be recovered based on source image distribution via a deep inversion technique. 
Using a membership inference attack strategy~\cite{nasr2019comprehensive, hu2022membership}, it is possible to infer whether a given sample exists or not in training dataset, thereby revealing private information.
The black-box SFUDA can help protect data privacy because only application programming interface (API) is accessible while detailed model weights are withheld, but it may suffer from performance degradation of cross-domain adaptation. 

\item Most white-box SFUDA methods assume that model architecture is shared between source and target domains, while the black-box SFUDA methods try to design task-specific target models for knowledge transfer.
Such flexible model design in black-box SFUDA methods is very useful for target users with low computation resources, since they can design more efficient and lightweight target models for domain adaptation. 

\item Black-box SFUDA methods neither require data synthesis nor model fine-tuning, which helps to accelerate the convergence process of model training. 
In contrast, white-box methods are usually computationally intensive and time-consuming.
For instance, it is reported that the computational cost of a black-box SFUDA method~\cite{zhang2022lightweight} is 0.83$s$ while that of two competing white-box methods are 3.17$s$~\cite{liang2020we} and 22.43$s$~\cite{ahmed2021unsupervised}, respectively, reflecting the computation efficiency of black-box SFUDA.
\end{itemize}

In summary, when using white-box and black-box SFUDA methods, we have to make a trade-off between obtaining better performance, protecting confidential information, and reducing computational and memory costs.

\subsection{Useful Strategies for Improved Generalizability}
To facilitate research practice in this field, we summarize several useful techniques that could be used to improve the generalizability of learning models for source-free unsupervised domain adaptation.

\subsubsection{Entropy Minimization Loss}
Most SFUDA methods utilize an entropy minimization loss~\cite{grandvalet2004semi} to reduce uncertainty of model predictions~\cite{liang2020we, bateson2022source, liu2022sourceEM, kothandaraman2021ss, wang2020tent, bateson2020source, liu2022unsupervised2, liu2022unsupervised1, liu2021adapting, fleuret2021uncertainty, hou2020source, xu2022source, kundu2021generalize}.
This simple yet effective strategy encourages the model to generate one-hot predictions for more confident 
learning.

\subsubsection{Diversity Enforcing Loss}
To prevent predicted labels from collapsing to categories with larger number of samples, many studies leverage a diversity enforcing loss to encourage diverse predictions over target domain~\cite{liang2020we, huang2022relative, wang2021give, roy2022uncertainty, liu2022sourceEM, liang2021umad, dong2021confident, tian2022source} .
The usual practice is to maximize the entropy of empirical label distribution over the batch-wise average of model predictions.

\subsubsection{Label Smoothing Technique}
In source-free adaptation studies, a pre-trained source model is generally obtained via training on labeled source data before adaptation stages.
Currently, many studies use a label smoothing technique~\cite{muller2019does, szegedy2016rethinking} to produce a robust source model~\cite{liang2020we, wang2022exploring, yang2021exploiting, ding2022proxymix, li2022sourcemd, chen2022source}.
This technique aims to transform original training labels from hard labels (\eg, 1) to soft labels (\eg, 0.95), which prevents the source model from being over-confident, helping enhance its generalization ability.
Also, the experiments have shown that label smoothing can encourage closer representations of training samples from the same category~\cite{muller2019does}.
With a more general and robust source model, it is likely to boost adaptation performance on target domain.

\subsubsection{Model Regularization}
Many regularization terms are utilized in existing SFUDA methods by incorporating some prior knowledge. 
For instance, an early learning regularization~\cite{qiu2021source, xu2022sourceelr, arpit2017closer} is used to prevent the model from over-fitting to label noise.
A stability regularization~\cite{li2020model, yan2021augmented, yang2022sourceself, xiong2022source} is leveraged to prevent parameters of the target model to deviate from those of the source model.
A local smoothness regularization~\cite{li2020model, ma2021semi} is used to encourage output consistency between the target model and its noise-perturbed counterpart, helping improve robustness of the target model.
A mixup regularization~\cite{liang2022dine, ding2022proxymix, guan2022polycentric, peng2022toward, kundu2022balancing} is used to enforce prediction consistency between original and augmented data, which can mitigate the negative influence of noisy labels.

\subsubsection{Confidence Thresholding}
Many studies leverage pseudo-labeling to train the target model in a self-supervised way.
Instead of utilizing a manually-designed threshold to identify reliable/confident pseudo-labels, a commonly used strategy is automatically learning the confidence threshold for reliable pseudo-label selection~\cite{li2022adaptive}.
To further tackle the class-imbalance problem, some studies~\cite{yang2022sourceself, kim2021domain, fleuret2021uncertainty, xu2022denoising, prabhu2021s4t} propose to learn dynamic threshold for each category, 
which provides a fair chance for categories with limited samples to generate pseudo-labels for self-training.

\section{Future Outlook}\label{sec_challenge_futurework}
\subsection{Multi-Source/Target Domain Adaptation}
To utilize diverse and rich information of multiple domains, a few studies~\cite{tian2021source, ahmed2021unsupervised, dong2021confident, shen2022benefits, han2022privacy} propose multi-source data-free adaptation to transfer source knowledge to the target domain.
Tian~\etal~\cite{tian2021source} introduce a sample transport learning method, but the proposed model is shallow, and thus cannot handle highly nonlinear feature extraction.
To tackle this problem, several deep learning based models~\cite{ahmed2021unsupervised, dong2021confident} are proposed. 
But they ignore the fact that the generated target pseudo-labels may be noisy, which may cause training bias when matching target domains with large domain gaps. 
The key to solving problems with multi-source domains is quantifying the transferability of different source models and utilizing their complementary information for promoting cross-domain adaptation.
Even several strategies are proposed (\eg, aggregation weight~\cite{ahmed2021unsupervised} and source-specific transferable perception~\cite{dong2021confident}), more explorations are encouraged to address the problem of negative transfer during cross-domain knowledge transfer. 

A few studies~\cite{peng2019federated, feng2021kd3a, kang2022privacy, song2020privacy, qin2022uncertainty} incorporate federated learning into domain adaptation scenarios.  
Federated learning~\cite{bonawitz2019towards, li2020federated, bonawitz2017practical} is a decentralized scheme to facilitate collaborative learning among multiple distributed clients without sharing training data or model parameters.
The constraint that prevents data and parameter transmission across different source domains is not required in multi-source-free domain adaptation.
For instance, \emph{federated adversarial domain adaptation (FADA)} introduced by Peng~\etal~\cite{peng2019federated} is among the first attempts to propose the concept of federated domain adaptation, which 
employs a dynamic attention mechanism to transfer knowledge from multi-source domains to an unlabeled target domain.
In this method, each source model needs to synchronize with the target domain after each training batch, resulting in huge computation costs and potential risk of privacy leakage~\cite{zhu2019deep}.
To tackle this problem, Feng~\etal~\cite{feng2021kd3a} introduces a consensus focus schema that greatly improves communication efficiency for decentralized domain adaptation.
Moreover, Song~\etal~\cite{song2020privacy} utilize a homomorphic encryption approach for privacy protection, and Qin~\etal~\cite{qin2022uncertainty} introduce a flexible uncertainty-aware strategy for reliable source selection.
However, current federated learning studies usually produce a common model for all clients without considering heterogeneity of data distribution of different clients.
Therefore, the common model cannot adapt to each client adaptively, which may affect adaptation performance.
It would be very interesting to investigate personalized federated learning~\cite{tan2022towards}, with which current or new clients can easily adapt to their own local dataset by performing a few optimization steps.
Besides, all the methods mentioned above require labeled data from multiple sources to train a federated model, inevitably increasing annotation costs.
Therefore, approaches that effectively exploit unlabeled data from multiple source domains in a decentralized way are urgently needed.

On the other hand, Yao~\etal~\cite{yao2022federated} and Shenaj~\etal~\cite{shenaj2022learning} have proposed several federated \emph{multi-target} domain adaptation strategies for transferring knowledge of a labeled source server to multiple unlabeled target clients.
More advanced techniques for federated multi-target domain adaptation are highly desirable, by considering computation and communication cost, annotation burden, and privacy protection of different target domains.

\subsection{Test-Time Domain Adaptation}
Most SFUDA approaches require pre-collected unlabeled target data for model training, termed ``training-time adaptation".
Test-time adaptation~\cite{wang2020tent, yazdanpanah2022visual, karani2021test, boudiaf2022parameter, iwasawa2021test} has been investigated by adapting the source model to the target domain during inference procedure only.
The advantages of test-time adaptation are mainly twofold: 
(1) The adaptation process does not need iterative training, which greatly improves computational efficiency, so the model can be easily deployed in an online manner.
(2) Without relying on target training data, test-time adaptation is expected to be well generalized to diverse target domains.
Even current studies have made promising achievements, there are still some problems worth exploring, listed as follows. 

Some studies~\cite{wang2020tent, ma2022test, you2021test} need to access batch-sized ($>$1) target samples during inference, which cannot handle scenarios where target samples arrive one-by-one sequentially.
Two studies~\cite{he2021autoencoder, he2020self} perform image-wise adaptation rather than batch-wise adaptation, but they cannot deal with cases with large distribution shift.
It is interesting to explore how to handle the scenarios where test instances come from continuous changeable domains in the future.
Additionally, the solutions on how to adaptively exploit test data can be further explored~\cite{yang2022dltta}, such as adjusting model weights dynamically based on sample discrepancy across domains.

\subsection{Open/Partial/Universal-Set Domain Adaptation}
This survey focuses on \emph{close-set} source-free domain adaptation, where the label space of source and target domains is consistent.
But practical scenarios are much more complicated when the category shift issue occurs across different domains.
There are three \emph{non-close-set} scenarios: (1) open-set ($C_s\subset C_t$) problems, (2) partial-set ($C_s \supset C_t$) problems, and (3) universal-set ($C_s \backslash C_t \neq \varnothing$ and $C_t \backslash C_s \neq \varnothing$, $C_s\subset C_t$, $C_s \supset C_t$) problems, where $C_s$ and $C_t$ denote the category label set for source and target domains, respectively.

Currently, only a few studies~\cite{kundu2020universal, kundu2020towards, zhao2022source} attempt to handle the category shift problem in source-free adaptation scenarios, including out-of-distribution data construction~\cite{kundu2020universal, kundu2020towards}, neighborhood clustering learning~\cite{saito2020universal}, uncertainty-based progressive learning~\cite{luo2022source}, and mutual information maximization~\cite{liang2021umad}.
The main idea behind these studies is to recognize out-of-source distribution samples and improve generalization ability of the source model.
However, the model performance of existing studies is not quite satisfactory due to the inaccessibility of valuable category-gap knowledge.
One possible solution is to adaptively learn a threshold instead of using a fixed one to determine the acceptance/rejection of each target sample as a ``known" category via some similarity measurement.
Moreover, some strategies used in non-source-free domain adaptation can also be borrowed, such as distribution weighted combining rule~\cite{xu2018deep}, category-invariant representation learning~\cite{lekhtman2021dilbert}, one-vs-all learning scheme~\cite{saito2021ovanet}, and global-local optimization~\cite{feng2021globally}.

\subsection{Flexible Target Model Design}
For black-box SFUDA methods, due to unavailability of structure and parameters of target model, one usually has to manually design a target model.
For instance, Liu~\etal~\cite{liu2022unsupervised1} choose a U-Net based framework as the target model for segmentation.
However, such manually designed architectures may be not suitable when adapting to the target domain.
It is expected that the automatic design of target models, \eg, using neural architecture search (NAS)~\cite{elsken2019neural, wistuba2019survey, lu2021neural}, helps improve the learning performance. 
Considering that NAS has recently become a popular strategy for searching proper network architectures in deep learning, we can integrate it into SFUDA scenarios to find more proper and efficient target models.
And how to balance the search space and search cost of network parameters can be further investigated.
Moreover, hyperparameters used in NAS (\eg, optimizer strategy, weight decay regularization) should be carefully considered since they also have a significant impact on network performance~\cite{ren2021comprehensive}.

\subsection{Cross-Modality Domain Adaptation}
Existing studies mainly focus on one single modality for domain adaptation, while a few studies perform cross-modality adaptation in source-free settings~\cite{hong2022source, ahmed2022cross}. 
For instance, for medical data analysis, the acquisition expense of computed tomography (CT) scans is generally less than that of magnetic resonance imaging (MRI) scans, hence it may greatly reduce annotation cost for a segmentation task when transferring a source model trained on CT images to MRI scans~\cite{hong2022source}. 
Moreover, in computer vision field, it would be promising to investigate cross-modality adaptation in the future, \eg, image$\rightarrow$video, which aims to achieve video recognition based on the source model trained on the image dataset. 
Also, how to effectively integrate multi-modality (\eg, image, sound, text, and video) data for domain adaptation in a source-free way is an interesting but 
not yet widely studied problem.

\subsection{Continual/Lifelong Domain Adaptation}
Most current studies focus on improving adaptation performance on the target domain while neglecting the performance on source domain, running the risk of catastrophic forgetting problems~\cite{kemker2018measuring}.
To address this issue, several solutions have been developed from different aspects, such as domain expansion~\cite{zhang2020unsupervised}, historical contrastive learning~\cite{huang2021model}, domain attention regularization~\cite{yang2021generalized}, and model perturbation~\cite{jing2022variational}, while there is still massive room for performance improvement. 
Inspired by continual/lifelong learning~\cite{kirkpatrick2017overcoming, li2017learning, lopez2017gradient, de2021continual}, continual domain adaptation has recently made great progress by investigating gradient regularization~\cite{tang2021gradient}, iterative neuron restoration~\cite{wang2022continual}, buffer sample mixture~\cite{taufique2021conda}, etc. 
The continual domain adaptation in source-free settings for mitigation of catastrophic forgetting remains an underdeveloped topic that can be further explored in the future.

\subsection{Semi-Supervised Domain Adaptation}
Source-free domain adaptation in semi-supervised settings (\ie, with a few labeled target data involved for model training) has also been explored in recent years~\cite{chidlovskii2016domain, kothandaraman2021ss, nelakurthi2018source}. 
It usually performs semi-supervised adaptation with the help of active learning~\cite{wang2022active, kothandaraman2022distilladapt}, model memorization revelation~\cite{yang2022revealing}, and consistency and diversity learning~\cite{wang2021learning}. 
There is still a lot of space for improvement with a limited number of labeled target samples, \eg, by fine-tuning the current source-free adaptation frameworks, but this is not the focus of this survey.

\section{Conclusion}\label{sec_conclusion}
In this paper, we provide a comprehensive review of recent progress in source-free unsupervised domain adaptation (SFUDA).
We classify existing SFUDA studies into white-box and black-box groups, and each group is further categorized based on different learning strategies.
The challenges of methods in each category and our insights are provided. 
We then compare white-box and black-box SFUDA methods, discuss effective techniques for improving adaptation performance, and summarize commonly used datasets. 
We finally discuss promising future research directions. 
It is worth noting that the research topic of source-free unsupervised domain adaptation is still in its early stages, and we hope this survey can spark new ideas and attract more researchers to advance this high-impact research field.

\bibliographystyle{IEEEtran}
\bibliography{SFUDA_Review}

\end{document}